\documentclass[lettersize,journal]{IEEEtran}
\usepackage{amsmath,amsfonts}
\usepackage{algorithmic}
\usepackage{algorithm}
\usepackage{verbatim}
\usepackage{array}
\usepackage{textcomp}
\usepackage{stfloats}
\usepackage{url}
\usepackage{verbatim}
\usepackage{graphicx}
\usepackage{cite}
\usepackage{booktabs}
\usepackage{hyperref}
\usepackage{enumerate}
\usepackage{MnSymbol}
\usepackage{ragged2e} 
\usepackage{booktabs,makecell, multirow, tabularx}
\usepackage{pifont}
\usepackage{subfigure}
\usepackage{xcolor}
\makeatletter

\newcommand{\Rmnum}[1]{\expandafter\@slowromancap\romannumeral #1@}
\makeatother
\hypersetup{hypertex=true,
colorlinks=true,
linkcolor=blue,
anchorcolor=blue,
citecolor=blue}

\begin{document}

\title{CLIP4VI-ReID: Learning Modality-shared Representations via CLIP Semantic Bridge for Visible-Infrared Person Re-identification}

\author{Xiaomei~Yang,~\IEEEmembership{}
        Xizhan~Gao,~\IEEEmembership{}
        Sijie~Niu,~\IEEEmembership{Member,~IEEE,}
        Fa~Zhu,~\IEEEmembership{}
        Guang~Feng,~\IEEEmembership{}
        Xiaofeng~Qu,~\IEEEmembership{}
        and~David Camacho ~\IEEEmembership{}
\thanks{The work was supported in part by the Shandong Provincial Natural Science Foundation under Grants Nos. ZR2025MS1004, ZR2025QC647, and in part by the National Natural Science Foundation of China under Grants Nos. 62471202, 62406143, 62101213.}
\thanks{Xiaomei Yang, Xizhan Gao, Sijie Niu, Guang Feng, Xiaofeng Qu are with the Shandong Key Laboratory of Ubiquitous Intelligent Computing, School of Information Science and Engineering, University of
Jinan, Jinan 250022, China (e-mail: ise\_gaoxz@ujn.edu.cn). \emph{(Corresponding author: Xizhan Gao.)}}
\thanks{Fa Zhu is with the College of Information Science and Technology \& Artificial Intelligence, Nanjing Forestry University, Nanjing 210037, China.}
\thanks{David Camacho is with the Computer Systems Engineering Department, Universidad Politécnica de Madrid, 28031, Calle Alan, Turing s/n, Madrid, Spain.}
}


\markboth{Journal of \LaTeX\ Class Files,~Vol.~?, No.~?, August~202?}%
{Shell \MakeLowercase{\textit{et al.}}: A Sample Article Using IEEEtran.cls for IEEE Journals}


\maketitle

\begin{abstract}
The core of visible-infrared person re-identification (VI-ReID) lies in learning modality-shared representations across different modalities. Given the significant modality gap between visible and infrared images, VI-ReID still remains a challenge. Recent research tends to use pre-trained CLIP to extract modality-shared representations, but they fail to consider differences in physical characteristics of natural images and infrared images when extracting visual features of infrared images and generating their text semantics. This causes the learned infrared text semantics to struggle to delicately describe the images, thereby limiting the performance of VI-ReID. In addition, after obtaining the text semantics, they do not make any adaptive adjustments, resulting in an increased risk of mismatches. To address these issues, we propose a novel CLIP-driven modality-shared representation learning network named CLIP4VI-ReID for VI-ReID task, which consists of Text Semantic Generation (TSG), Infrared Feature Embedding (IFE), and High-level Semantic Alignment (HSA). Specifically, considering the huge gap in the physical characteristics between natural images and infrared images, the TSG is designed to generate text semantics only for visible images, thereby enabling preliminary visible-text modality alignment. Then, the IFE is proposed to rectify the feature embeddings of infrared images using the generated text semantics. This process injects id-related semantics into the shared image encoder, enhancing its adaptability to the infrared modality. Besides, with text serving as a bridge, it enables indirect visible-infrared modality alignment. Finally, the HSA is established to refine the high-level semantic alignment. This process ensures that the fine-tuned text semantics only contain id-related information, thereby achieving more accurate cross-modal alignment and enhancing the  discriminability of the learned modal-shared representations. Extensive experimental results demonstrate that the proposed CLIP4VI-ReID achieves superior performance than other state-of-the-art methods on some widely used VI-ReID datasets. The code will be available at https://github.com/y0406/CLIP4VI-ReID.
﻿
\end{abstract}


\section{Introduction}
\label{sec:1}

\begin{figure}[t]
\centering
\includegraphics[width=3.5 in]{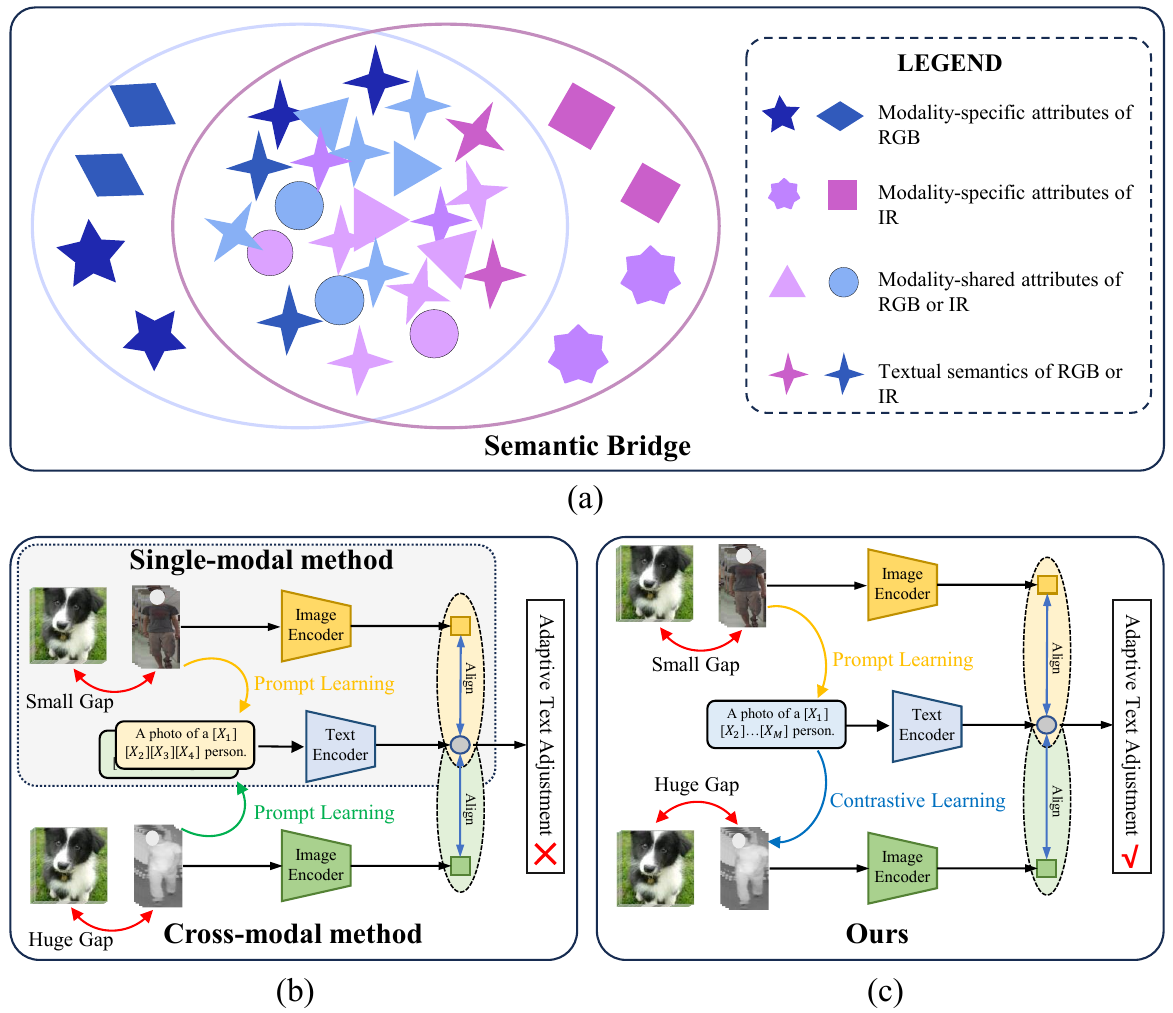}
\caption{The motivation of the paper: (a) Visible (RGB) images and infrared (IR) images share the same high-level semantics. Therefore, the text semantics can serve as a bridge to mitigate the modality gap. (b) Existing single-modality methods struggle to handle VI-ReID tasks and exhibit weak feature representation for infrared images. Besides, due to the huge gap between natural images and infrared images, existing cross-modality methods struggle to capture the fine-grained details of infrared objects, and may introduce semantic noise, thereby affecting their performance. Furthermore, they did not adaptively adjust the obtained text semantics. (c) In contrast, our CLIP4VI-ReID model proposes a coarse-to-fine RGB-IR cross-modal alignment scheme, i.e., it first generates text semantics for visible images, then uses the text semantics as a bridge to correct the embedding of infrared images. Finally, it adaptively adjusts the obtained text semantics and refines the high-level semantic alignment among the three modalities.  }
\label{fig:1}
\end{figure}

\IEEEPARstart{P}{erson} re-identification (ReID) is a critical computer vision and computational intelligence task aimed at retrieving query objects from large-scale gallery databases captured by non-overlapping cameras \cite{Zhu2025AE-Net,ye2021deep}. This task plays an essential role in enhancing public security, for instance, with the aid of intelligence surveillance systems, ReID enables the tracking of criminal suspects and the locating of missing persons. In recent years, driven by the increasing demand for computational intelligence and behavior analysis, this field has garnered significant global attention and achieved remarkable advancements in computer vision research \cite{Yan2024IsISILA}. However, most existing ReID models focus on visible-visible person re-identification, which performs effectively under favorable lighting conditions \cite{dong2023erasing,ye2021deep,zhang2022person,Singh2025TROPE,Zhang2022MFEA}. Nevertheless, their performance often deteriorates under low-light or nighttime conditions, limiting their applicability in modern systems requiring all-weather monitoring. Infrared (IR) cameras offer a viable solution by capturing thermal imaging information even in complete darkness, 
providing an alternative that is not constrained by ambient light. Given that most modern social surveillance systems are equipped with dual-mode cameras capable of automatically switching modes based on environmental lighting, it is imperative to investigate visible-infrared cross-modal person ReID (VI-ReID) technology. VI-ReID enables pedestrian matching between visible and infrared images across varying lighting conditions, thereby supporting 24/7 monitoring requirements.

Different from single-modal person ReID, VI-ReID focuses on retrieving images from a gallery of infrared or visible images that match the identity of a given visible or infrared query. In this case, 
the modalities of the gallery and query are different. As a result, researchers encounter significant challenges, including inter-modality differences and intra-modality variations. Inter-modality differences are primarily manifested in the substantial discrepancies between visible (RGB) and IR images, which encompass aspects such as color, texture, and brightness. RGB images are distinguished by their rich color information, whereas infrared images predominantly capture thermal radiation data. This inherent distinction introduces additional complexity to the person re-identification task \cite{zhang2022dual}. 
Intra-modality variations refers to the appearance differences among images of the same person captured under the same modality at different times, locations, or camera viewpoints. 
These differences can stem from various factors, such as changes in illumination, occlusion, pose variation, inconsistent resolution, complex backgrounds, which may become even more pronounced in a cross-modal tasks.

To address these challenges, researchers have proposed many cross-modal VI-ReID methods \cite{lu2023tri}, which can be grouped into modality-compensated methods \cite{zhang2022fmcnet,cui2024dma,lu2023learning,liu2022revisiting}  and modality-shared methods \cite{qiu2024high,fang2023visible,zhao2022spatial,hu2022adversarial,Chan2025DFHFL,wei2023dual,chen2023identity,liang2023cross,feng2022visible}. Modality-compensated methods first use some generative models such as GAN \cite{liu2020enhancing,xiang2019cross} to generate missing modalities from existing modalities to reduce inter-modality differences. Then, they extract discriminative features from the original and generated modalities to deal with the intra-modality variation problem. However, the presence of noise interference hinders the training stability, making it challenging to ensure the quality of generated images \cite{yu2025clip}. On the contrary, modality-shared methods first map two modality images into a shared feature space, and then, they use some loss functions \cite{wang2025learning} or decoupling modules \cite{wang2025DIRL} to enhance the discriminative ability of shared features. 
Although existing methods have shown certain effectiveness in using solely images to train models, such manner leads to the lack of high-level semantic information of the visual content learned by the model \cite{li2023clip}.

Recently, the visual language learning paradigm has attracted much attention for its powerful ability to extract semantically rich visual features \cite{Tang2025CGSKTLM}. The representative Contrastive Language Image Pre-training (CLIP) model \cite{radford2021learning} has achieved remarkable success in numerous downstream vision tasks \cite{bahng2022exploring,zhou2022learning}. And it has also been successfully used in the person Re-ID field, including single-modality ReID methods \cite{yu2024tf, li2023clip} and cross-modality ReID methods \cite{yu2025clip, Hu2025CLIPMC}. As shown in the upper left part of Fig. \ref{fig:1}(b), single-modality ReID methods (such as CLIP-ReID \cite{li2023clip} ) transfer CLIP to the image- or video-based person ReID tasks and show that combining visual information with corresponding linguistic description through CLIP can make the model perceive high-level semantic information related to the target pedestrian. However, these methods are designed for RGB data and cannot effectively handle cross-modality ReID tasks. Therefore, how to apply CLIP to cross-modality ReID remains a major challenge. A key to this challenge is that although traditional modality-shared features (e.g., texture and structure) have certain universality, they may fail to fully capture the unique information of pedestrian identities. Fortunately, as illustrated in Fig. \ref{fig:1}(a), features from different modalities share the same high-level semantics---a property that can compensate for the limitations of low-level shared features. Thus, the textual modality can serve as a bridge for cross-modal alignment to mitigate the modality gap. Based on this insight, some CLIP-based cross-modality ReID methods have emerged, which use CLIP as the backbone and leverages semantic information as a bridge for aligning the visual representations of visible and infrared images. As shown in Fig. \ref{fig:1}(b), despite achieving promising results, existing cross-modality ReID methods still face the following limitations during its construction process: Firstly, they directly use the visual encoder of the pre-trained CLIP model to generate text semantics for IR images, implicitly assuming that the pre-trained CLIP can effectively align IR images and text. However, due to the significant differences in physical characteristics between natural images and IR images, this assumption may not hold \cite{cao2025irgpt}. CLIP is pre-trained on large-scale datasets of natural images and corresponding text, where the images are primarily visible-light images rich in color, texture, and semantic information. In contrast, IR images mainly reflect temperature differences and exhibit significant differences in imaging principles and feature distributions compared to visible-light images. As a result, directly using the pre-trained CLIP model to generate text semantics for IR images may introduce noise\footnote{To illustrate this, we provide several representative failure cases in Fig. \ref{failure}, where IR-derived text semantics fail to capture fine-grained ID-related attributes, whereas RGB-derived semantics remain more semantically consistent.}, thereby affecting the Re-ID performance. In addition, they also need to design additional networks to fuse textual semantics, which increases the number of model parameters. Secondly, after obtaining the text semantics, they does not perform any adaptive adjustment. If the generated text semantics are inaccurate, they can directly impact the subsequent results. 

In order to address the aforementioned issues and inspired by the CLIP-ReID method, as shown in Fig. \ref{fig:1}(c), we propose the CLIP4VI-ReID network for VI-ReID task. More specifically, as shown in Fig. \ref{fig:2}, our method consists of three stages: Text Semantic Generation (TSG), Infrared Feature Embedding (IFE), and High-level Semantic Alignment (HSA). Firstly, the TSG is used to generate text tokens for RGB modality images. This stage leverages the prompt learning mechanism and the cross-modal alignment capabilities inherent in pre-trained CLIP to enhance the text encoder's ability to capture fine-grained visual information from RGB images. As a result, the text encoder can generate text semantic descriptions of the RGB modality, which serve as the foundation for subsequent 
stages. Secondly, the IFE is utilized to correct the embedding of infrared modality features. This stage injects id-related semantics from the textual descriptions into the shared image encoder to enhance its 
adaptability to the IR modality. Subsequently, leveraging the text semantics as a bridge, it achieves an initial alignment between the RGB and IR modality features. Finally, the HSA is exploited to refine the high-level semantic alignment among the three modalities. This stage performs fine-tuning on the text encoder to obtain text semantics containing only id-related information and on the modality-specific image encoders to acquire more accurate modality-specific features. This process enables more precise RGB-IR cross-modal feature alignment and enhances the discriminability of modality-shared features.


Our main contributions are summarized as follows:
\begin{enumerate}[i)]
  \item We propose a novel CLIP-driven modality-shared representation learning network named CLIP4VI-ReID for visible-infrared person ReID task. Unlike previous VI-ReID approaches, we design a three-stream network architecture based on the CLIP model and propose a three-stage learning strategy to gradually achieve coarse-to-fine RGB-IR cross-modal feature alignment. As a new modality-shared feature learning method, it can effectively enhance the visible-infrared cross-modal alignment performance and improve the discriminative ability of modality-shared features.
  \item The CLIP4VI-ReID model adopts a staged learning strategy, with each stage having specific objectives and optimized loss functions. Staged learning helps the model gradually learn more fine-grained feature representations. In addition, by generating text semantics only for visible images, it can avoid noise interference caused by differences in physical characteristics during infrared text generation. Meanwhile, it can also simplify the model design and reduce the difficulty of learning.
  \item The proposed network incorporates semantic information into the visual representation, which can improve the discriminative ability of modality-shared features. We validate the performance of our proposed method on two widely used VI-ReID benchmark datasets. The experimental results show that the proposed method outperforms state-of-the-art (SOTA) approaches in almost all cases, demonstrating its effectiveness and superiority.
\end{enumerate}


\section{Related Work}
\label{sec:2}
In this section, we will first review some related works, including single-modal person ReID methods and visible-infrared person ReID methods. Then, we will review vision-language pre-trained models and their latest research progress in person ReID field.

\subsection{Single-modal Person ReID}
Person re-identification \cite{Wang2025NPSSL} is a computer vision technology aimed at determining whether a specific individual appears in an image. It is widely regarded as a sub-problem of image retrieval. Typically, person ReID refers to single-modal person re-identification (SmReID), which involves retrieving visible images of a monitored pedestrian across devices given a visible image of that person. The primary challenge in SmReID lies in addressing the degradation of recognition performance caused by factors such as viewpoint variation, background interference, pose changes, and occlusion. To tackle this challenge, researchers have proposed various methods. Some of these methods focus on designing new network architectures to learn more discriminative features. For example, He et al. \cite{he2021transreid} introduced a Transformer-based image re-identification network (TransReID) to overcome the limitations of CNNs. Gao et al. \cite{gao2024part} proposed a teacher-student decoding (TSD) framework leveraging a Transformer decoder and human parsing for occluded person re-identification. Zhu et al. \cite{zhu2019progressive} presented a novel pose transfer method for generative adversarial networks to transform the pose of a given person to the target pose. Other methods focus more on employing different loss functions to enhance the distance metrics. For instance, Hermans et al. \cite{hermans2017defense} proposed to use triplet loss for person ReID task. Luo et al. \cite{luo2019strong} integrated label smoothing into the cross-entropy loss function. Ye et al. \cite{ye2016person} achieved precise search results by aggregating multiple ranked lists. Sarfraz et al. \cite{sarfraz2018pose} directly computed the average distance of extended neighbors as the final measurement. Despite the robust extraction of pedestrian information achieved by various algorithms in single-modal person ReID under favorable lighting conditions, practical applications still face challenges, particularly in nighttime or low-light environments where the performance of these methods may degrade due to difficulties in extracting effective features.

\subsection{Visible-Infrared Person ReID}
Cross-modal person ReID, specifically visible-infrared person ReID, aims to retrieve an infrared image with the same identity given a visible person image, and vice versa. Compared to SmReID, 
VI-ReID is more challenging due to significant modality differences. Existing VI-ReID methods can be broadly categorized into modality-shared methods and modality-compensated methods. Modality-shared methods aim to extract discriminative cross-modal features while discarding modality-specific features to address both cross-modal and intra-modal variations in VI-ReID. For instance, Fang et al. \cite{fang2023visible} proposed the SASR framework, which effectively mitigated the impact of pedestrian motion and enhanced the uniqueness of feature representations. Chen et al. \cite{chen2022structure} proposed a SPOT network for learning semantic-level shared cross-modal representations. Feng et al. \cite{feng2022visible} introduced the CMIT framework, which integrated the strengths of CNNs and Transformers by exchanging information between modalities. Zhao et al. \cite{zhao2022spatial} proposed the THC loss, which learned discriminative feature representations by balancing cross-modal and intra-modal distances of centers. Modality-compensated methods aim to compensate for missing modality-specific information in existing modalities, thereby reducing modality discrepancies. For example, Li et al. \cite{li2020infrared} introduced an auxiliary X mode to bridge the gap between visible and infrared modalities. Liu et al. \cite{liu2022revisiting} proposed a two-stage modality enhancement network for VI-ReID, which incorporated a high-quality cross-modal image generator, a feature-level fusion module, and an assisted learning module to effectively address modality differences. Zhang et al. \cite{zhang2022fmcnet} proposed compensating at the feature level rather than the image level. Despite achieving some success, existing VI-ReID methods have not fully utilized the high-level semantic information contained in pedestrian images, and there is still room for further improvement in their performance.

\subsection{Vision-Language Model-based Person ReID}  
As an emerging research direction, the multimodal \cite{Tang2025DaC, Tang2022LAgPF} and vision-language learning is progressively transforming the research paradigm in the field of computer vision. CLIP \cite{radford2021learning} is a representative vision-language learning model, which establishes a connection between natural language and visual content through the contrastive learning of image-text pairs. Up to now, CLIP has achieved remarkable success in numerous downstream vision tasks \cite{zhou2022learning}, including person ReID. For instance, Li et al. \cite{li2023clip} pioneered the CLIP-ReID model, which combined learnable prompts with pedestrian identity language descriptions during training and utilized natural language supervision to extract semantically rich image features. Yu et al. \cite{yu2024tf} proposed the TF-CLIP framework, which transfered CLIP to video-based person ReID task by replacing text features with CLIP-memory. Meanwhile, researchers have also extended CLIP to the VI-ReID field. Hu et al. \cite{Hu2025CLIPMC} proposed the CLIP-based Modality Compensation (CLIP-MC) method, which proposed an instance text prompt generation strategy to reduce the difference between image and text encoders, and it was the first CLIP work in modality-specific information compensation VI-ReID field. Yu et al. \cite{yu2025clip} proposed the CSDN method, which used CLIP as the backbone and leveraged semantic information as a bridge for aligning the visual representations of visible and infrared images. These methods have achieved promising results; however, the full potential of CLIP in VI-ReID remains underexplored. In light of this, we innovatively propose a novel CLIP-driven modality-shared representation learning network for visible-infrared person ReID task.

\begin{figure*}[t!]
\centering
\includegraphics[width=1\textwidth]{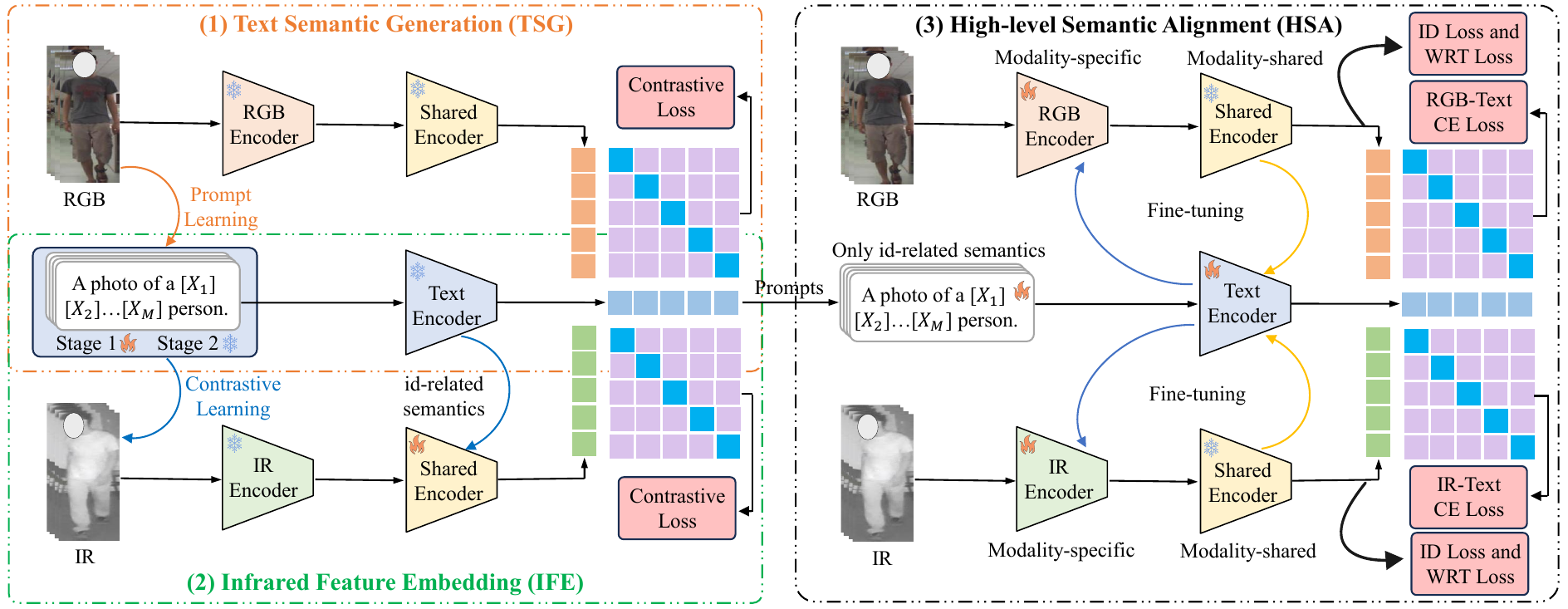} %
\caption{Overview of the proposed CLIP4VI-ReID method. CLIP4VI-ReID is a three-stream model, and its learning process includes three stages: TSG, IFE and HSA.  TSG is first utilized to generate text semantics for RGB-modality images, then IFE uses the text semantics as a bridge to correct the embedding of IR-modality images. Finally, HSA adaptively adjusts the obtained text semantics and refines the high-level semantic alignment among the three modalities.}
\label{fig:2}
\end{figure*}


\section{Methodology}
\label{sec:3}
In view of the significant modality gap between visible and infrared images, current researches tend to use shared CNN/ViT networks or pre-trained CLIP model to extract modality-shared features for cross-modal alignment. However, the former relies exclusively on images for network training, thereby making it difficult to perceive high-level semantic information, which is essential for mitigating modality gaps. The latter directly employs pre-trained image encoders to learn visual features for RGB and IR images, ignoring the huge gap between natural images (used in CLIP's training) and IR images, resulting in inaccurate semantic information learning. Therefore, we proposed the CLIP4VI-ReID framework, which takes text semantics as a bridge and employs a three-stage learning strategy to achieve coarse-to-fine RGB-IR cross-modal feature alignment (as shown in Fig. \ref{fig4}). Fig. \ref{fig:2} depicts the overall framework of CLIP4VI-ReID, which includes TSG, IFE and HSA. 

\subsection{Preliminaries}
VI-ReID aims to retrieve images with the same identity as the provided visible (infrared) query from a gallery of infrared (visible) images; therefore, the cross-modal VI-ReID dataset can be expressed as $D = \left\{ \left( X_i^v, X_i^r, y_i \right) \mid i = 1, 2, \ldots, N \right\}$, where  $X^v_i = \left\{ x^v_{(i,j)} \mid j = 1, 2, \ldots, M_i^v \right\}$ represents the image set of the $i^{th}$ identity under the visible modality, where $M_i^v$ is the number of images in $X^v_i$. $X_i^r = \left\{ x^r_{(i,k)} \mid k = 1, 2, \ldots, M_i^r \right\}$ represents the image set of the $i^{th}$ identity under the infrared modality, where $M_i^r$ is the number of images in $X_i^r$. $y_i$ is the label of the $i^{th}$ pedestrian identity, used to identify images of the same pedestrian.

As shown in Fig. \ref{fig:2}, to achieve the objectives of this paper, we design a three-stream network architecture based on the CLIP model. Specifically, we utilize the image encoder of pre-trained CLIP to construct the RGB and IR branches, and we use the text encoder $E_t(\cdot)$ of CLIP to construct our text branch. Besides, to adapt to the VI-ReID task, we modified the two image branches (i.e., the RGB and IR branches). That is, we used the first blocks of the ResNet-based CLIP image encoder to extract modality-specific features and the last four blocks to extract modality-shared features. Thus, we obtained the modality-specific RGB encoder $E_v(\cdot)$, the modality-specific IR encoder $E_r(\cdot)$, and the modality-shared encoder $E_s(\cdot)$.

\subsection{Text Semantic Generation}
For CLIP to reach its full potential for high-level semantic perception, a key prerequisite is that natural language descriptions that match images must be available. However, such textual descriptions are often missing in pedestrian image datasets, which greatly limits the application of CLIP in VI-ReID tasks. Inspired by CLIP-ReID \cite{li2023clip}, CSDN \cite{yu2025clip} employs prompt learning to generate semantic descriptions for RGB and IR modalities respectively. However, due to the significant modality gap between natural images and IR images, the learned IR text semantics struggle to delicately describe the images, thereby limiting the performance ceiling of ReID. To overcome this limitation, we propose the TSG to generate text semantics only for RGB-modality images, which can promote strong coupling between text semantics and RGB images, ensuring more fine-grained textual descriptions are generated.

To generate semantically rich text descriptions for RGB images, we design a structured learnable textual description $T_i$: ``a photo of a $[X_1^i]$, $[X_2^i]$, \ldots , $[X_M^i]$ person'' for the $i^{th}$ pedestrian identity, where each $[X_m^i]$ ($m=1, \ldots , M$) is a learnable text token whose dimensions are consistent with the word embedding, $M$ represents the total number of learnable text tokens.
As shown in Fig. \ref{fig:2}, we first use the text encoder $E_t(\cdot)$ to extract the text features 
\begin{equation}\label{1}
f^t_i = E_t(T_i), i=1, 2, \cdots, N.
\end{equation}

Then, we use the image encoder (i.e., RGB encoder $E_v(\cdot)$ and shared encoder $E_s(\cdot)$) to extract the RGB-modality image features:
\begin{equation}\label{2}
f_k^v = E_s (E_v (x^v_k)), k=1, 2, \cdots, n_b,     
\end{equation}
where $n_b$ denotes the batch size, $x_k^v$ represents the $k^{th}$ image in the batch. In order to ensure an accurate matching between $T_i$ and visible image $x_k^{v}$, we adopt a bidirectional contrastive loss function, including the contrastive loss $L_{v2t}^1$ from visible image to text and the contrastive loss $L_{t2v}^1$ from text to visible image. This bidirectional contrastive loss function can not only ensure the semantic consistency between the image and the text, but also optimize the learnable text tokens through backpropagation, so as to generate more semantically representative text descriptions. In this process, we freeze the parameters of $E_v(\cdot)$, $E_s(\cdot)$, and $E_t(\cdot)$, and only optimize the text token $[X_m^i]$. By minimizing the loss of $L_{v2t}^{1}$ and $L_{t2v}^{1}$, the gradient is backpropagated through $E_t(\cdot)$ to optimize $[X_1^i], [X_2^i], \ldots, [X_M^i] $, thus making full use of the powerful semantic extraction capability of $E_t(\cdot)$. The loss function at this stage is:
\begin{equation}\label{eq:1}
L_{v2t}^1 = -\frac{1}{n_b}\sum_{k=1}^{n_b} \log \frac{\exp(s(f_k^v, f_k^t))}{\sum_{j=1}^{n_b} \exp(s(f_k^v, f_j^t))},
\end{equation}
\begin{equation}\label{eq:2}
L_{t2v}^1 = -\frac{1}{n_b}\sum_{k=1}^{n_b} \frac{1}{|P(y_k)|} \sum_{p_k \in P(y_k)} \log \frac{\exp(s(f_{p_k}^v, f_{y_k}^t))}{\sum_{j=1}^{n_b} \exp(s(f_j^v, f_{y_k}^t))},
\end{equation}
where $f^v_k$ represents the feature of the $k^{th}$ RGB image in the batch, $f^t_k$ is the text feature corresponding to $f^v_k$, $y_k$ represent the label of $k^{th}$ image in the batch, $P(y_k)$ represents an index set composed of sample indices with identity $y_k$. $|P(y_k)|$ denotes the number of samples in $P(y_k)$. Finally, the total loss function is formulated as:
\begin{equation}\label{eq:3}
L_{1} =L_{v2t}^1 + L_{t2v}^1 .
\end{equation}

\begin{figure*}[t!]
	\centering
	\includegraphics[width=0.8\textwidth]{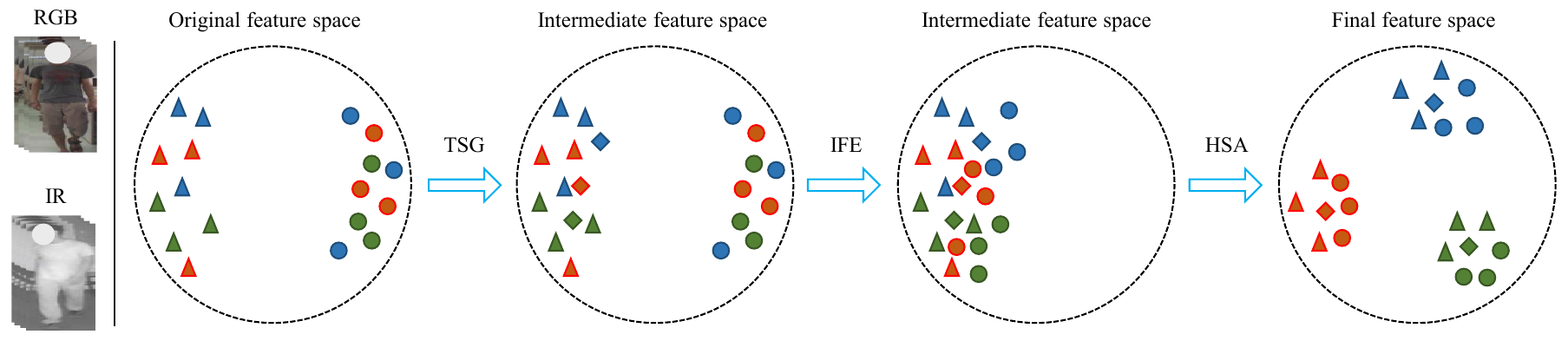} %
	\caption{Illustration of the three-stage learning process of the proposed CLIP4VI-ReID method. Triangles and circles represent RGB-modality features and IR-modality features, respectively, while diamonds represent text-modality features. Different colors represent different pedestrian categories.}
	\label{fig4}
\end{figure*}

Through this subsection, we obtain the initial text semantics $T_1, T_2, \cdots, T_N$ for all $N$ ids. As shown in Fig. \ref{fig4}, these text semantics have achieved preliminary alignment with RGB images. To further improve the computational efficiency, we save the text features $f_1^t, f_2^t, \cdots, f_N^t$ corresponding to these semantics. This strategy not only improves the computational efficiency, but also provides a stable feature base for the subsequent stages of optimization.

\subsection{Infrared Feature Embedding}
As mentioned in the previous subsection, due to the significant differences in physical characteristics between natural images and IR images, CLIP cannot effectively extract features from IR modality images. Directly using a frozen CLIP image encoder to extract features from IR images may introduce noisy. Fortunately, RGB and IR images share the same text semantics. Therefore, in this subsection, we utilize the text semantics to correct the feature embeddings of the infrared modality images and achieve preliminary RGB-IR feature alignment.

The text semantics generated by TSG consist of two parts: high-level semantics that describe shared features (such as id information) and low-level semantics that describe modality-specific features (such as color, texture, etc.). The high-level semantics are learned by the shared encoder, while the low-level semantics are learned by the RGB encoder. In this stage, we aim to inject the id-related semantics (i.e., high-level semantics) from the textual descriptions into the IR branch to enhance its adaptability to the IR modality images. Therefore, we only train the shared encoder in this branch to accurately extract the modality-shared features from IR images. More specifically, we freeze the encoders $E_r(\cdot)$ and $E_t(\cdot)$, and only optimize the shared encoder $E_s(\cdot)$, so as to lay a foundation for the subsequent extraction of modal shared features. 
In order to ensure the accurate correspondence between $T_i$ and infrared image $x_{k}^{r}$, we also adopt a bidirectional contrastive loss function, including the IR-to-text contrastive loss $L_{r2t}^2$ and the text-to-IR contrastive loss $L_{t2r}^2$. This bidirectional contrastive loss function can not only ensure the semantic consistency between the image and the text, but also optimize the parameters of the shared encoder $E_s(\cdot)$ through back propagation, 
so as to generate more accurate modality-shared feature representations. The loss function at this stage is:
\begin{equation}\label{eq:4}
L_{r2t}^2 = -\frac{1}{n_b}\sum_{k=1}^{n_b} \log \frac{\exp(s(f_k^r, f_k^t))}{\sum_{j=1}^{n_b} \exp(s(f_k^r, f_j^t))},
\end{equation}
\begin{equation}\label{eq:5}
L_{t2r}^2 = -\frac{1}{n_b}\sum_{k=1}^{n_b} \frac{1}{|P(y_k)|} \sum_{p_k \in P(y_k)} \log \frac{\exp(s(f_{p_k}^r, f_{y_k}^t))}{\sum_{j=1}^{n_b} \exp(s(f_j^r, f_{y_k}^t))}.
\end{equation}
where $f^r_k = E_s(E_r(x^r_k))$ represents the feature of the $k^{th}$ IR image in the batch.

Finally, the total loss function is formulated as:
\begin{equation}\label{eq:6}
L_{2} =L_{r2t}^2 + L_{t2r}^2 .
\end{equation}

Through this subsection, we obtain the corrected infrared feature embeddings, as shown in Fig. \ref{fig4}, these feature embeddings have achieved preliminary alignment with RGB images and text semantics.

\subsection{High-level Semantic Alignment}
After successfully implementing TSG and IFE, our method is already equipped with a finely optimized shared encoder and has achieved preliminary RGB-IR feature alignment. However, as shown in Fig. \ref{fig4}, CLIP4VI-ReID still faces the issue of weak discriminative ability. We guess this is because the generated text semantics contain excessive RGB modality-specific information, which undermines the discriminative ability of the modality-shared features. Theoretically, optimal text semantics should exclusively contain id-related (i.e., modality-shared) information. Additionally, the pre-trained modality-specific encoders (i.e., the RGB encoder and IR encoder) still have limitations in extracting modality-specific details, which indirectly affects the discriminative ability of modality-shared features. To address the aforementioned issues, we propose the HSA to refine the high-level semantic alignment among the three modalities and enhance the discriminative ability of the modality-shared features.

Specifically, to achieve the above objectives, we use the following RGB-to-text cross entropy (CE) loss and IR-to-text cross entropy loss to simultaneously optimize the three encoders $E_v(\cdot)$, $E_r(\cdot)$, $E_t(\cdot)$, and the learnable text tokens, while keeping the already optimized shared encoder fixed.
\begin{equation}\label{eq:11}
L_{v2tce}^{3} = -\frac{1}{n_b}\sum_{k=1}^{n_b} q_k \log \frac{\exp(s(f_k^v, f_{y_k}^t))}{\sum_{y_a=1}^{N} \exp(s(f_k^v, f_{y_a}^t))},
\end{equation}

\begin{equation}\label{eq:12}
  L_{r2tce}^{3} = -\frac{1}{n_b}\sum_{k=1}^{n_b} q_k \log \frac{\exp(s(f_k^r, f_{y_k}^t))}{\sum_{y_a=1}^{N} \exp(s(f_k^r, f_{y_a}^t))},
\end{equation}
where $n_v = n_r = n_b/2$, $q_k$ is the one-hot vector of identity label $y_k$.

To further improve the discriminative ability  of our model, we adopt identity (id) loss and weighted regularization triplet (wrt) loss to optimize the visible feature $f^v$ and infrared feature $f^r$, ensuring that these features are highly discriminative.
\begin{equation}\label{eq:8}
L_{id} = -\frac{1}{n_b}\sum_{i=1}^{n_b} q_i \log(W(f_i)), 
\end{equation}
where $ f_i = f_i^v or f_i^r $ is the output feature of $E_s(\cdot)$; $W$ denotes the identity classifier.
\begin{equation}\label{eq:9}
L_{wrt} = \frac{1}{n_b} \sum_{i=1}^{n_b} \log(1 + \exp(\sum_{i,j} w_{ij}^p d_{ij}^p - \sum_{i,k} w_{ik}^n d_{ik}^n)),
\end{equation}

\begin{equation}\label{eq:10}
w_{ij}^p = \frac{\exp(d_{ij}^p)}{\sum_{d_{ij}^p \in P_i} \exp(d_{ij}^p)}   ,
\ w_{ik}^n = \frac{\exp(-d_{ik}^n)}{\sum_{d_{ik}^n \in N_i} \exp(-d_{ik}^n)},
\end{equation}
where $j$ and $k$ are the indices of the positive and negative samples corresponding to anchor $x_i$ (here $x_i$ can be $x^v_i$ or $x^r_i$); $P_i$ and $N_i$ respectively represent the positive and negative sample sets corresponding to $x_i$ in a batch; $d_{ij}^p = \|f_i - f_j\|_2$ 
and $d_{ik}^n = \|f_i - f_k\|_2$ denote the Euclidean distance of the positive and negative sample pairs.

Finally, the loss function of HSA is given by
\begin{equation}\label{eq:13}
L_{hsa} = \lambda_1 L_{v2tce}^{3} +  \lambda_2 L_{r2tce}^{3} + L_{id} + L_{wrt},
\end{equation}
where $\lambda_1$, $\lambda_2$ are the balance hyper-parameters. As shown in Fig. \ref{fig4}, through this subsection, we not only ensure the discrimination of features within modalities, but also realize the sharing of high-level semantic information between modalities, thus achieving finer cross-modal feature alignment and significantly improving the accuracy and robustness of cross-modal person re-identification.


\begin{algorithm}[t]
\caption{CLIP4VI-ReID}
\label{alg:1}
\textbf{Input:} Sample set $ \mathcal{D} = \{ X_i^v, X_i^r, y_i \}_{i=1}^N$, pre-trained CLIP, hyper-parameters $\lambda_1, \lambda_2$, number of iterations $S_1, S_2$, and $S_3$. \\
\textbf{Output:} The trained visual encoders $E_v$, $E_r$, and $E_s$. 
\begin{algorithmic} [1]
\STATE Initialize text semantics $T_i$ and classifier $W$;
\STATE \textbf{Stage 1:} Training the TSG 
\FOR {$iter=1,2,\cdots,S_1$}
\STATE Update text semantics $T_i$ by minimizing Eq. (\ref{eq:3});
\ENDFOR
\STATE \textbf{Stage 2:} Training the IFE
\FOR {$iter=1,2,\cdots,S_2$}
\STATE Update $E_s$ by minimizing Eq. (\ref{eq:6});
\ENDFOR
\STATE \textbf{Stage 3:} Training the HSA
\FOR {$iter=1,2,\cdots,S_3$}
\STATE Update $E_v$, $E_r$, $E_t$, $T_i$ and $W$ by minimizing Eq. (\ref{eq:13}).
\ENDFOR
\end{algorithmic}
\end{algorithm}


\subsection{Training and Inference}
Our proposed method divides the whole training process into three stages. Initially, in the first stage, we optimize the learnable text token of the visible image under the constraints of Eq. (5), which is responsible for generating the text description $T_i, i=1, 2, \cdots, N$ corresponding to the visible image. Then, in the second stage, we impose Eq. (11) to train the shared encoder $E_s(\cdot)$ for obtaining corrected IR feature embeddings, laying a solid foundation for the subsequent extraction of cross-modal shared features. Finally, the third stage involves training modality-specific visual encoder $E_v(\cdot)$, $E_r(\cdot)$, an identity classifier $W$ and a text encoder $E_t(\cdot)$ using the constraints specified in Eq. (14), facilitating discriminative and modal invariance learning of visual representations. In particular, it is important to emphasize that in the inference phase, we only rely on the features generated by the visual encoder and compute the cosine similarity for identity recognition. Such a design makes other components other than the visual encoder no longer necessary during inference, thus ensuring the efficiency and convenience of our designed network (CLIP4VI-ReID) in practical applications. In addition, Algorithm \ref{alg:1} outlines the detailed steps of the learning flow, which makes the whole process more transparent and easy to understand.


\begin{figure}[!t]
\centering
\includegraphics[width=3.5 in]{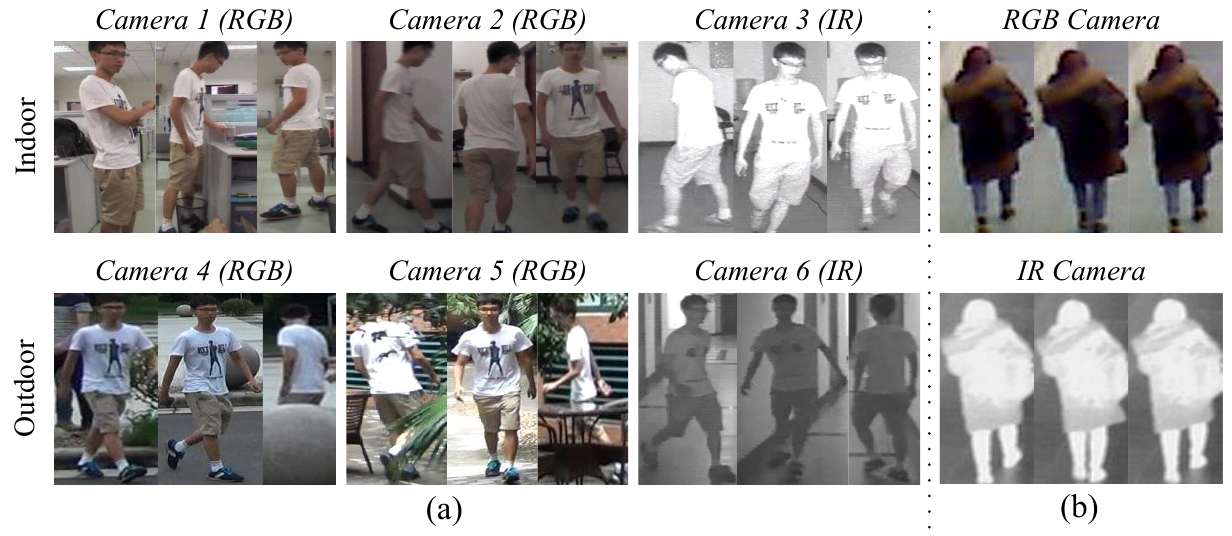}
\caption{Examples from (a) SYSU-MM01 dataset and (b) RegDB dataset.}
\label{dataset}
\end{figure}

\section{Experiments}
\label{sec:4}
\subsection{Datasets}
To demonstrate the effectiveness of the proposed CLIP4VI-ReID, we conduct comparison experiments on two publicly available VI-ReID datasets, that is, SYSU-MM01 \cite{wu2017rgb} and RegDB \cite{nguyen2017person}.

\subsubsection{\textbf{SYSU-MM01\cite{wu2017rgb}}} This is the first large-scale VI-ReID dataset, which contains 15,792 infrared and 287,628 visible images of 491 pedestrian identities. These images are captured by ``Camera 1'' - ``Camera 6'' under all-search mode and indoor-search mode, respectively. Each mode consists of two settings: single-shot and multi-shot. In our experiments, we use 96 identities as the testing set, which includes 3,803 infrared images for querying, and 301 randomly selected visible images for gallery. Fig. \ref{dataset}(a) displays some example images from SYSU-MM01 dataset.

\subsubsection{\textbf{RegDB\cite{nguyen2017person}}} RegDB is another popular VI-ReID dataset, which contains 4,120 infrared and 4,120 visible images of 412 pedestrian identities. In our experiments, 206 identities are used to construct training set, and the remaining 206 identities are utilized as testing set. In testing phase, there are two evaluation settings: visible-to-infrared and infrared-to-visible. All experiments are repeated 10 times and the average results are recorded. Fig. \ref{dataset}(b) displays some example images of RegDB dataset.

\subsection{Experiment Settings}
\subsubsection{\textbf{Evaluation Protocol}} According to the existing VI-ReID methods, we use cumulative matching characteristic (CMC), mean Average Precision (mAP) and mINP as our evaluation metrics.

\begin{table*}[!t]
\renewcommand\arraystretch{1.35}
\caption{Comparison results with the SOTA methods on SYSU-MM01 dataset. The bold and underline represent the optimal and sub-optimal results, respectively. }
\label{result_sysu}
\vspace{-1mm}
\centering
\resizebox{\textwidth}{!}{%
 \begin{tabular}{c| c | c c c c c | c c c c c | c c c c c| c c c c c }
    \hline
    \multirow{4}*{Methods} & 
    \multirow{4}{*}{Reference} & 
    \multicolumn{20}{c}{SYSU-MM01} \\
   \cline{3-22}
    & & \multicolumn{10}{c|}{All-Search} & \multicolumn{10}{c}{Indoor-Search}\\
   \cline{3-22}
    & & \multicolumn{5}{c|}{Single-Shot} & \multicolumn{5}{c|}{Multi-Shot} & \multicolumn{5}{c|}{Single-Shot} & \multicolumn{5}{c}{Multi-Shot} \\
   \cline{3-22}
    & &                   Rank-1  & Rank-10 & Rank-20 & mAP  & mINP   & Rank-1  & Rank-10 & Rank-20 & mAP  & mINP   & Rank-1  & Rank-10 & Rank-20 & mAP  & mINP  & Rank-1  & Rank-10 & Rank-20 & mAP  & mINP \\
    \hline 
    $\ast$ AlignGAN\cite{wang2019rgb} & ICCV'19  
    & 42.40    & 85.0  & 93.7   & 40.70   & -       & 51.50    & 89.4  & 95.7   & 33.90   & -      & 45.90    &87.6   &94.4    & 54.30   & -       & 57.10    & 92.7  &97.4    & 45.30   & -    \\
    $\ast$ XIV\cite{li2020infrared}       & AAAI'20   
    &49.92     &89.79  &95.96 &50.73     & -       & -        & -     & -      & -       & -     & -        & -     & -      & -       & -      & -        & -     & -      & -       & -    \\
    DDAG\cite{ye2020dynamic}& ECCV’20  
    &54.75 &90.39     &95.81      &53.02    & 39.62  & -        & -    & -    & -       & -       &61.02 &94.06 &98.41 & 67.98  & 62.61  & -        & -    & -    & -       & -   \\
    AGW\cite{ye2021deep}& TPAMI'21 
    & 47.50    & 84.3 & 92.1    & 47.65   & 35.30   & -        & - & - & -     & -      & 54.17    & 91.1 &95.9 & 62.97  &59.23  & -        & - & - & -      & -   \\
    MPANet\cite{wu2021discover}    & CVPR'21  
    & 70.5     &96.2 &98.8      & 68.2    &         & 75.5     & 97.7 &99.4 & 62.9  & -      & 76.7     & 98.2 &99.5 & 80.9   & -     & 84.2     &99.6 &99.9 &75.1    & -     \\  
    SFANet\cite{liu2021sfanet}    & TNNLS'21 
    & 65.74    & 92.98 & 97.05 & 60.83   & -  & -        & - & -& -     & -  & 71.60    & 96.60 & 99.45 & 80.05  & -   & -        & - & -& -      & -    \\
    \hline 
    MID\cite{huang2022modality}       & AAAI'22  
    & 60.27    & 92.90 & - & 59.40   & -       & -        & - & - & -     & -      & 64.86    &  96.12 & - & 70.12  & -     & -        & - & -  & -      & -      \\
    $\ast$ FMCNet\cite{zhang2022fmcnet}    & CVPR'22  
    & 66.34    & - & - & 62.51   & -       & -        & - & -& -     & -     & 68.15    & - & -& 74.09  & -     & -        & - & - & -      & -    \\
    SPOT\cite{chen2022structure}      & TIP'22  
     & 65.34   & 92.73 & 97.04 & 62.25   & 48.86  & -       & - & - & -     & -     & 69.42   & 96.22 & 99.12 & 74.63  & 70.48 & -       & - & - & -      & -   \\
     $\ast$ TSME\cite{liu2022revisiting}      & TCSVT'22 
    & 64.23     & 95.19 & 98.73 & 61.21    & -      & 70.34     & 96.75 & 99.26  & 54.36  & -     & 64.80     & 96.92 & 99.31 & 71.53   & -      & 76.83     & 98.84 & 99.89 & 65.02   & -    \\
    \hline
    PMT\cite{lu2023learning}       & AAAI'23  
    &67.53     &95.36  &98.64   &64.98     &\underline{51.86}   & -        & -     & -      & -        & -      &71.66     &96.73  &99.25   &76.52     &72.74  & -        & -     & -      & -        & -    \\
    TOPLight\cite{yu2023toplight}  & CVPR'23  
    &66.76     &96.23     &98.70 &64.01 &50.18  & -        & -    & - & -     & -      & 72.89    & 97.93  &99.28  & 76.70  & 71.95 & -        & - & -  & -      & -    \\
    CMTR\cite{liang2023cross}      & TMM'23   
    & 65.45    & 94.47 & 98.16 & 62.90   & -      & 71.99    & 96.37 & 99.09 & 57.07 & -    & 71.46    & 97.16 & 99.22 & 76.67  & -    & 80.00    & 98.53 & 99.71 & 69.49  & -   \\
    CMIT\cite{feng2022visible}      & TMM'23  
    & 70.94    & 94.93 & 96.37 & 65.51   & -    & -        & - & - & -     & -    & 73.28    & 95.20 & 99.43 & 77.18  & -   & -	       & - & - &-       & -   \\
    DSCNET\cite{zhang2022dual}    & TIFS'23  
    & 73.89    & 96.27 & 98.84 & 69.47   & -   & -        & - & - & -     & -    & 79.35    & 98.32 & \underline{99.77} & 82.65  & -   & -        & - & - & -      & -    \\
    DFLN-ViT\cite{zhao2022spatial}  & TMM'23  
    & 59.84    & 92.49 & 97.20 & 57.70   & -      & -        & - & - & -     & -     & 62.13    & 94.83 & 98.24 & 69.03  & -   & -        & - & - & -      & -    \\
    \hline 
    MIP\cite{wu2024enhancing}       & ICMR'24 
    & 70.84    & - & - & 66.41   & -     & 74.68    & - & -& 58.49 & -     & 78.80    & - & - & 79.92  & -    & 82.29    & - & -& 70.65  & -    \\
    TMD\cite{lu2023tri}       & TMM'24   
    & 73.92    & 96.29 & 98.76 & 67.76   & 51.52  & -        & - & - & -     & -    & 81.16    &\underline{98.87} & 99.68 & 78.88  & \underline{73.35} & -        & - & - & -      & -     \\
    CSDN\cite{yu2025clip}     &  TMM'24        
    & 75.2     & 96.6   & 98.8   &71.8 & -  & \textbf{80.6} & \underline{98.3} & \underline{99.7} & \textbf{66.3} & - &82.0 & 98.7 & 99.5 & \underline{85.0}   & - & \underline{88.5} & \underline{99.6} & \underline{99.9} & \underline{80.4} & -  \\
      \hline
    DMPF\cite{lu2024disentangling}    & TNNLS'25  & \textbf{76.41}  & 96.40  & 98.66 & 71.55 & -        & - & - & -     & -  & - & \underline{82.29}   & 98.64  & 99.73  & 84.94 & -        & - & - & -     & -  & - \\ 
    CSCL\cite{liu2024cross}            & TMM'25   & \underline{75.72} & \underline{97.14} & \underline{99.18} & \underline{72.08} & -        & - & - & -     & -  & - & 80.80  & 98.14  & 99.42 & 83.56   & -        & - & - & -     & -  & - \\ 
        
     \hline
    \textbf{OURS} & - & 75.54        & \textbf{97.33} & \textbf{99.41}      & \textbf{72.55}    & \textbf{60.36} 
                      & \underline{79.82}     & \textbf{98.31} & \textbf{99.73}      & \underline{63.89} & \textbf{22.31} 
                      & \textbf{83.98}        & \textbf{99.23} & \textbf{99.94}      & \textbf{86.31}    & \textbf{82.94}
                      & \textbf{90.34}     & \textbf{99.93} & \textbf{100}      & \textbf{80.46} & \textbf{49.06}

    \\
    \hline
    \end{tabular}
}
\end{table*}

\begin{table}[!t]
\renewcommand\arraystretch{1.5}
\caption{Comparison results with the SOTA methods on RegDB dataset. The bold and underline represent the optimal and sub-optimal results, respectively. }
\label{result_regdb}
\vspace{-1mm}
\centering
\resizebox{0.485\textwidth}{!}{%
 \begin{tabular}{c| c | c c c c c | c c c c c}
    \hline
    \multirow{4}*{Methods} & 
    \multirow{4}{*}{Reference} & 
    \multicolumn{10}{c}{\multirow{2}{*}{RegDB}} \\
    & & \multicolumn{5}{c}{} & \multicolumn{5}{c}{} \\
   \cline{3-12}
    & & \multicolumn{5}{c|}{Infrared to Visible} & \multicolumn{5}{c}{Visible to Infrared} \\
   \cline{3-12}
    & &   Rank-1  & Rank-10 & Rank-20 & mAP  & mINP   & Rank-1  & Rank-10  & Rank-20 & mAP  & mINP \\
    \hline
    $\ast$ AlignGAN\cite{wang2019rgb}  & ICCV'19     & 56.30  & -       & -       & 53.40   & -       & 57.90   & -        & -       & 53.60  & -  \\ 
    $\ast$ XIV\cite{li2020infrared}       & AAAI'20     & -  & -       & -       & -   & -       &62.21   &83.13   &91.72  &60.18  & -  \\ 
    DDAG\cite{ye2020dynamic}      & ECCV’20     & 68.06  & 85.15  &90.31      & 61.80   & 48.62   & 69.34   &86.19 & 91.49         & 63.46  & 49.24\\
    AGW\cite{ye2021deep}       & TPAMI'21    & 70.49  & -       & -       & 65.90   & -       & 70.05   & -       & -        & 66.37  & -\\
    MPANet\cite{wu2021discover}     & CVPR'21     & 82.8   & -       & -       & 80.7    & -       & 83.7    & -       & -        & 80.9   & -   \\   
    SFANet\cite{liu2021sfanet}     & TNNLS'21    & 70.15  & 85.24   &89.27      & 63.77   & -       & 76.31   &  91.02 & 94.27       & 68.00  & - \\ 
    \hline 
    MID\cite{huang2022modality}       & AAAI'22     & 84.29  &  93.44       & -       & 81.41   & -       & 87.45   & 95.73       & -        & 84.85  & - \\ 
    $\ast$ FMCNet\cite{zhang2022fmcnet}     & CVPR'22     & 88.38  & -       & -       & 83.86   & -       & 89.12   & -       & -        & 84.43  & - \\ 
    SPOT\cite{chen2022structure}       & TIP'22      & 79.37  & 92.79       & 96.01       & 72.26   & 56.06   & 80.35   & 93.48       & 96.44        & 72.46  & 56.19\\ 
   $\ast$  TSME\cite{liu2022revisiting}      & TCSVT'22    & 84.4   & 96.39       & 98.20       & 75.7    & -       & 87.3    & 97.10       & 98.90        & 76.9   & -\\ 
    \hline
    PMT\cite{lu2023learning}        & AAAI'23     & 84.16  & -       & -       & 75.13   & -       & 84.83   & -       & -        & 76.55  & - \\ 
    DEEN\cite{zhang2023diverse}        & CVPR'23     & 89.5   & 96.8       & 98.4       & 83.4    & -       & 91.1    & 97.8       & 98.9        & 85.1   & - \\ 
    TOPLight\cite{yu2023toplight}  & CVPR'23     & 80.65  & 92.81       & 96.32       & 75.91   & 59.26   & 85.51   & 94.99       & 96.70        & 79.95  &63.85\\ 
    CMTR\cite{liang2023cross}       & TMM'23      & 84.92  & -       & -       & 80.79   & -       & 88.11   & -       & -        & 81.66  & - \\ 
    CMIT\cite{feng2022visible}      & TMM'23      & 84.55  & 93.72       & 95.83       & 83.64   & -       & 88.78   & 94.76       & 97.04        & \underline{88.49}  & -\\ 
    DSCNET\cite{zhang2022dual}    & TIFS'23     & 83.50  & -       & -       & 75.19   & -       & 85.39   & -       & -        & 77.30  & - \\ 
    
    \hline 
    MIP\cite{wu2024enhancing}        & ICMR'24     & \textbf{92.38}   & -       & -        &  \underline{85.99}  & -  & 91.26  & -       & -       & 85.90   & -      \\ 
   CSDN\cite{yu2025clip}            &   TMM'24  & 88.2 & 95.1 & 96.6 & 82.8 & - & 89.0 & 96.1 & 97.9 & 84.7 & - \\

    TMD\cite{lu2023tri}        & TMM'24      & 87.38  & 96.07       & 97.96       & 81.28   &  \underline{68.49}    &  \underline{92.96}   &  \underline{98.74}       &  \underline{99.37}        & 84.34  &  \underline{72.79}  \\ 
    \hline
     DMPF\cite{lu2024disentangling}    & TNNLS'25  & 88.88  &\underline{97.62}  & \underline{98.83} & 81.86 & - & 88.83 & 97.38 & 98.35 & 81.02 & - \\
      CSCL\cite{liu2024cross}         & TMM'25    & 89.66 & - & - &85.07 & - & 92.16  & - & - & 84.27 & - \\

    \hline
    \textbf{OURS} & - 
    & \underline{92.28}    &\textbf{98.01}      & \textbf{99.27}                   & \textbf{86.58} & \textbf{74.51} 
    & \textbf{94.51}    & \textbf{98.88}       & \textbf{99.56}                   &  \textbf{88.55} & \textbf{77.26} 
    \\
    \hline
    \end{tabular}
}
\end{table}
    

\subsubsection{\textbf{Experimental Details}} We carefully adopt PyTorch as a deep learning framework to implement our CLIP4VI-ReID, and choose to conduct all experiments on an excellent GTX 3090 GPU (24GB). 
To ensure the consistency of the input data, all input images are resized to a uniform size of 288$\times$144 pixels, and we apply a series of standard image data augmentation techniques, including random flipping, random padding, and random cropping, to enhance the generalization ability of the model. In the first two stages of training, we first train the learnable text tokens for the visible image and subsequently train the shared encoder for the infrared image, and each stage performs 120 training epochs. The initial learning rate during training is set to $1 \times 10^{-5}$ at epoch 0. As the epoch increases, the learning rate gradually increases until it reaches $3 \times 10^{-4}$ in the 5th epoch. After that, the learning rate starts to decrease according to the cosine annealing strategy. Subsequently, in the third stage, we train the infrared encoder, the visible encoder, the text encoder, and the visible text tokens and classifiers for a total of 180 epochs. In this phase, the learning rate starts from $3 \times 10^{-6}$, increases linearly to 0.0003 in the first 10 epochs, and decays by a ratio of 0.1 in the 60th and 100th epoch, respectively. The batch size is set to 32 on the SYSU-MM01 dataset and 16 on the RegDB dataset. Besides, we adopt the Adam optimizer \cite{kingma2014adam}, and set the hyperparameters to $\lambda_1 = 0.05$ and $\lambda_2 = 0.05$, respectively.

\subsection{Comparison with SOTA Methods}
In order to demonstrate the effectiveness and competitiveness of our approach, we present a comprehensive comparison of CLIP4VI-ReID against SOTA methods on two widely recognized VI-ReID benchmarks. The results are summarized in Table \ref{result_sysu}  and Table \ref{result_regdb}.

\subsubsection{\textbf{SYSU-MM01}}
We first conduct experiments on the SYSU-MM01 dataset. As illustrated in Table \ref{result_sysu}, the proposed CLIP4VI-ReID consistently outperforms state-of-the-art methods across most cases. 
Specifically, under all-search (single-shot) setting, our CLIP4VI-ReID method achieves the optimal Rank-10, Rank-20, mAP and mINP of 97.33\%, 99.41\%, 72.55\% and 60.36\%, respectively. Compared to sub-optimal method CSCL, CLIP4VI-ReID reduces the errors of by 6.64\%, 28.05\%, and 1.68\%, respectively. Compared to most relevant method CSDN, CLIP4VI-ReID reduces the errors of  Rank-1, Rank-10, Rank-20 and mAP by 1.37\%, 21.47\%, 50.83\% and 2.66\%, respectively. Under all-search (multi-shot) setting,  CLIP4VI-ReID method achieves the optimal Rank-10, Rank-20 and mINP of 98.31\%, 99.73\% and 22.31\%, respectively, and it also achieves the sub-optimal Rank-1 and mAP. Under indoor-search single-shot and multi-shot settings, our CLIP4VI-ReID achieves optimal results in all cases. To be specific, under indoor-search (single-shot) setting, CLIP4VI-ReID achieves the optimal Rank-1, Rank-10, Rank-20, mAP and mINP of 83.98\%, 99.23\%, 99.94\%, 86.31\% and 82.94\%, respectively, which reduces the error of sub-optimal method TMD by 14.97\%, 31.86\%, 81.25\%, 35.18\% and 35.98\%, and reduces the error of the most relevant method CSDN by 11\%, 40.77\%, 88\%, 8.73\% and 100\%. Under indoor-search (multi-shot) setting, CLIP4VI-ReID achieves the optimal Rank-1, Rank-10, Rank-20, mAP and mINP of 90.34\%, 99.93\%, 100\%, 80.46\% and 49.06\%, respectively. Compared to sub-optimal method CSDN,  CLIP4VI-ReID reduces the errors of  Rank-1, Rank-10, Rank-20 and mAP by 16\%, 42.86\%, 100\% and 0.31\%, respectively. In addition, we also found that compared with traditional VI-ReID methods, the CLIP-based models (i.e., CSDN and CLIP4VI-ReID) achieve much better performance, which fully demonstrates the ability of CLIP in high-level semantic perception. Compared with generative-based methods (e.g., AlignGAN, XIV, FMCNet and TSME), CLIP4VI-ReID obtains much higher results in all settings. Compared with CSDN, our CLIP4VI-ReID achieves better results in most cases, which demonstrates the effectiveness of the proposed method and is consistent with the previous theoretical analysis.

\begin{table}[!t]
\renewcommand\arraystretch{1.03}
\begin{center}
\caption{The effects of our proposed components performed on SYSU-MM01 dataset. }
\label{sysu_ablation}
\vspace{-3mm}
\resizebox{0.485\textwidth}{!}{ 
    \begin{tabular}{c| c c c | c c | c c }
    \hline
    \multirow{2}*{Methods} & 
    \multicolumn{3}{c|}{Components} & \multicolumn{2}{c|}{All-Search} & \multicolumn{2}{c}{Indoor-Search} \\
    \cline{2-8}
    & TSG & IFE & HSA & Rank-1 & mAP & Rank-1 & mAP  \\
    \hline
     1(Base)     &           &           &           & 69.60 & 66.08 & 77.01 & 78.81\\
    \hline
    2           & \ding{52} &           &           & 69.84 & 66.39 & 77.23 & 79.07\\
    3           &           &           & \ding{52} & 70.71 & 67.85 & 77.27 & 79.43\\ 
    4           & \ding{52} &           & \ding{52} & 73.17 & 70.38 & 81.59 & 84.24\\ 
    5           & \ding{52} & \ding{52} &           & 71.26 & 69.43 & 79.17 & 82.86\\
    \hline
    \textbf{6(Full)}     & \ding{52} & \ding{52} & \ding{52} & \textbf{75.54} & \textbf{72.55} & \textbf{83.98} & \textbf{86.31}\\
    \hline
    \end{tabular}
    }
\end{center}
\end{table}

\begin{table}[!t]
\renewcommand\arraystretch{1.06}
\begin{center}
\caption{The effects of our proposed components performed on RegDB dataset.}
\label{regdb_ablation}
\vspace{-3mm}
\resizebox{0.485\textwidth}{!}{ 
    \begin{tabular}{c| c c c | c c | c c }
    \hline
    \multirow{2}*{Methods} & 
    \multicolumn{3}{c|}{Components} & \multicolumn{2}{c|}{Visible to Infrared} & \multicolumn{2}{c}{Infrared to Visible}  \\
    \cline{2-8}
    & TSG & IFE & HSA & Rank-1 & mAP & Rank-1 & mAP  \\
    \hline
    1(Base)     &           &           &           & 86.84 & 80.24 & 84.76 & 79.74\\ 
    \hline
    2           & \ding{52} &           &           & 87.11 & 80.41 & 84.92 & 80.08\\ 
    3           &           &           & \ding{52} & 88.50 & 81.56 & 86.31 & 80.89\\  
    4           & \ding{52} &           & \ding{52} & 91.41 & 83.97 & 89.22 & 81.98\\  
    5           & \ding{52} & \ding{52} &           & 89.66 & 85.26 & 85.48 & 82.47\\  
    \hline
    \textbf{6(Full)}     & \ding{52} & \ding{52} & \ding{52} & \textbf{94.51} & \textbf{88.55} & \textbf{92.28} & \textbf{86.58} \\
    \hline
    \end{tabular}
    }
\end{center}
\end{table}

\begin{table}[!t]
\renewcommand\arraystretch{1.06}
\begin{center}
\caption{The influence of learnable text tokens on RegDB dataset.
}
\label{learnable_token}
\vspace{-3mm}
\resizebox{0.485\textwidth}{!}{ 
    \begin{tabular}{c|c| c c | c c }
    \hline
    \multirow{2}*{Methods} 
   & \multirow{2}*{Prompt Tokens} & \multicolumn{2}{c|}{Visible to Infrared} & \multicolumn{2}{c}{Infrared to Visible}  \\
    \cline{3-6}
   & & Rank-1 & mAP & Rank-1 & mAP  \\
    \hline 
    A photo of [class]  &0   & 91.84 & 84.63 & 89.63 & 82.14\\ 
    \hline
    \text{Learnable prompts}   &2  & 92.38 & 86.18 & 90.52 & 84.31 \\
    \hline
    \text{Learnable prompts}   &4  & \textbf{94.51} & \textbf{88.55} & \textbf{92.28} & \textbf{86.58} \\
    \hline
    \text{Learnable prompts}   &6  & 93.16 & 87.40 & 91.19 & 85.89 \\
    \hline
    \text{Learnable prompts}   &8  & 93.59 & 87.37 & 92.07 & 86.30 \\
    \hline
    \end{tabular}
    }
\end{center} 
\end{table}

\subsubsection{\textbf{RegDB}}
To provide a more comprehensive evaluation of our approach, we perform comparison experiments with state-of-the-art methods on the RegDB dataset, and the experimental results are reported in Table \ref{result_regdb}. It can be seen from this table that in almost all cases the proposed CLIP4VI-ReID method achieves the best performance (i.e., achieves the optimal results 9 times), while TMD method achieves the sub-optimal performance (i.e., achieves the sub-optimal results 5 times). More specifically, under infrared-to-visible setting, our method achieves the Rank-1, Rank-10, Rank-20, mAP and mINP of 92.28\%, 98.01\%, 99.27\%, 86.58\% and 74.51\%, which reduces the errors of MIP by 4.21\% (mAP), of TMD by 38.83\% (Rank-1), 49.36\% (Rank-10), 64.22\% (Rank-20), 28.31\% (mAP), 19.11\% (mINP), of CSDN by 34.58\% (Rank-1), 59.39\% (Rank-10), 78.53\% (Rank-20), 21.98\% (mAP). Under visible-to-infrared setting, we find again that our method achieves the best performance in all cases. To be specific, CLIP4VI-ReID boasts 94.51\% Rank-1 accuracy, 98.88\% Rank-10, 99.56\% Rank-20, 88.55\% mAP and 77.26\% mINP, reducing the errors of the sub-optimal method TMD by 22.02\% (Rank-1), 11.11\% (Rank-10), 30.16\% (Rank-20), 26.88\% (mAP), 16.43\% (mINP). Compared with generative-based methods (e.g., AlignGAN, XIV, FMCNet and TSME), we find again that the proposed CLIP4VI-ReID obtains much higher results in all cases. Besides, we also found that the proposed CLIP4VI-ReID outperforms CSDN in all cases, which indicates that the motivation of this paper is correct.



\begin{figure}[!t]
\centering
\includegraphics[width=3.5 in]{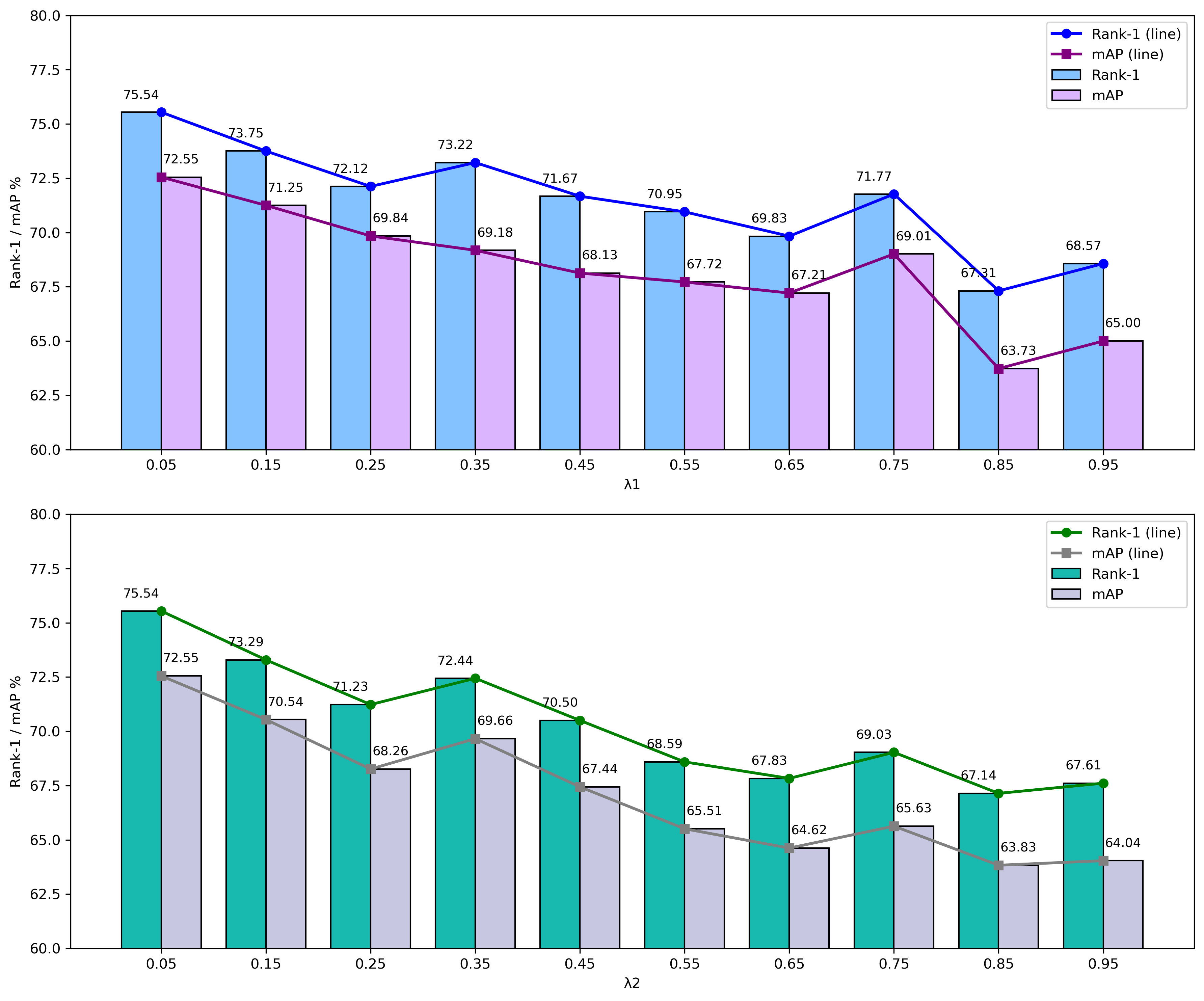}
\caption{The impact of different parameters $\lambda_1$ and $\lambda_2$ on the SYSU-MM01 dataset.}
\label{lambda1}
\end{figure}

\begin{figure}[!t]
\centering
\includegraphics[width=3.5 in]{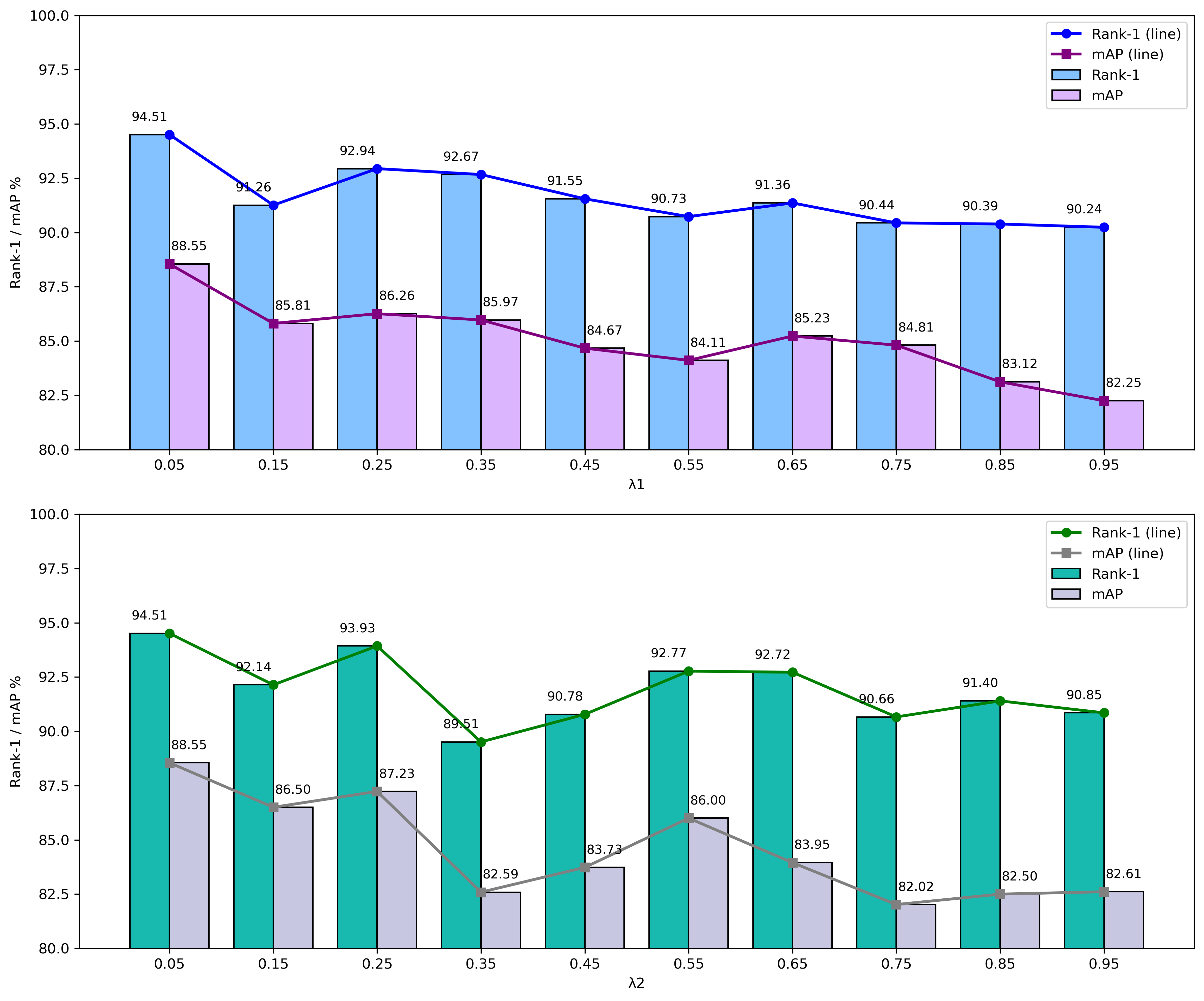}
\caption{The impact of different parameters $\lambda_1$ and $\lambda_2$ on the RegDB dataset.}
\label{lambda2}
\end{figure}

\begin{figure*}[t]
\centering
  \subfigure[Pre-trained CLIP (baseline)]{
    \includegraphics[width=0.22\textwidth]{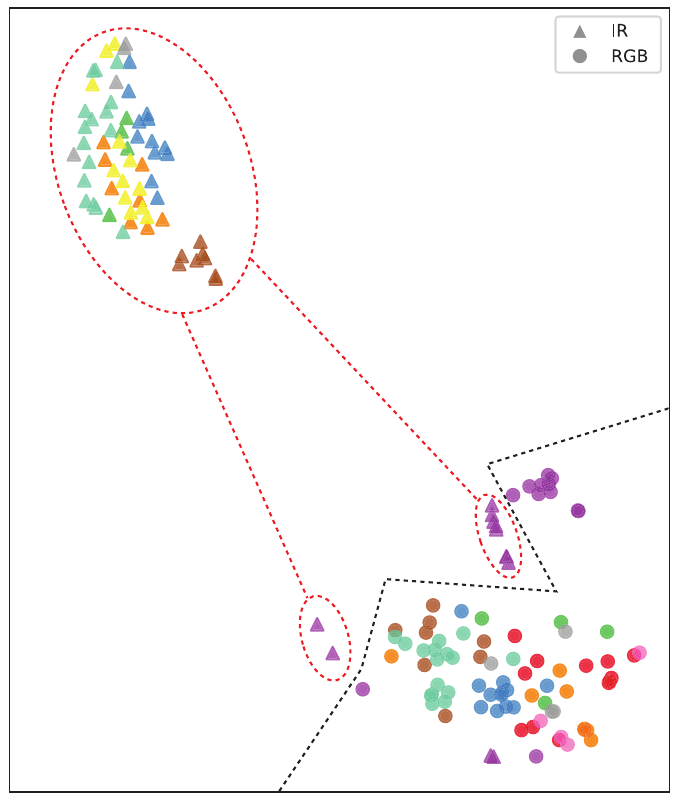}}
  \subfigure[TSG]{
    \includegraphics[width=0.22\textwidth]{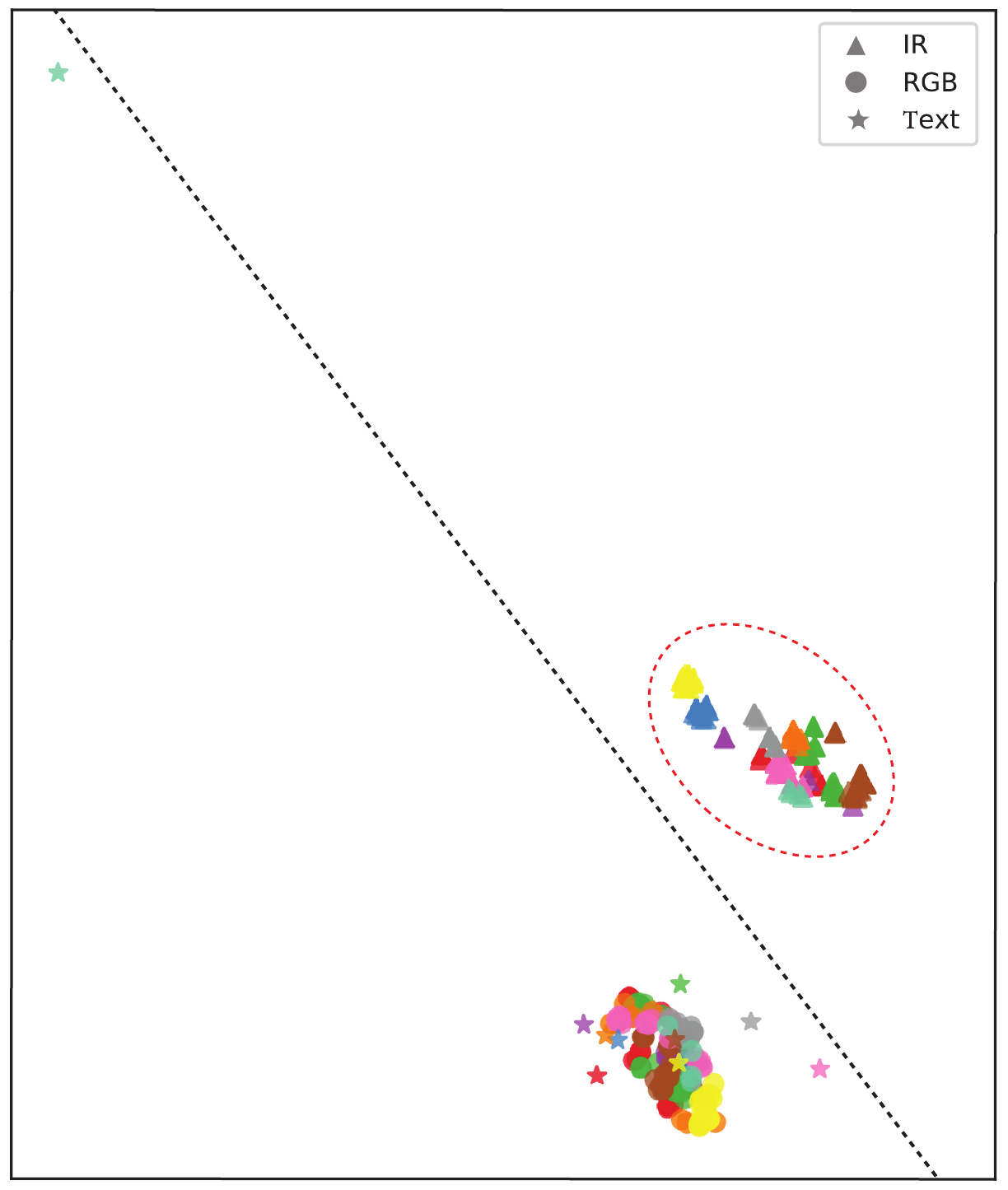}}
  \subfigure[IFE]{
    \includegraphics[width=0.22\textwidth]{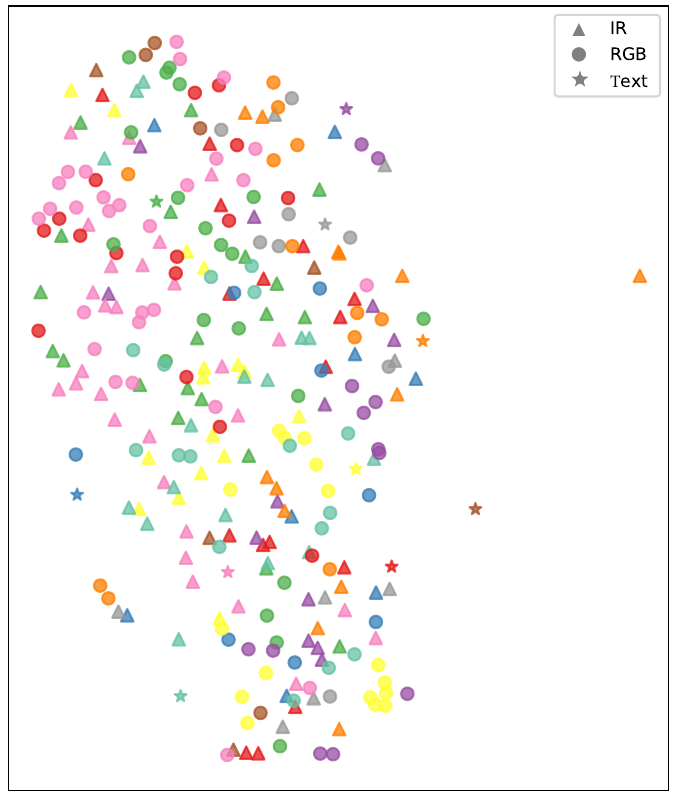}}
  \subfigure[HSA (CLIP4VI-ReID)]{
    \includegraphics[width=0.22\textwidth]{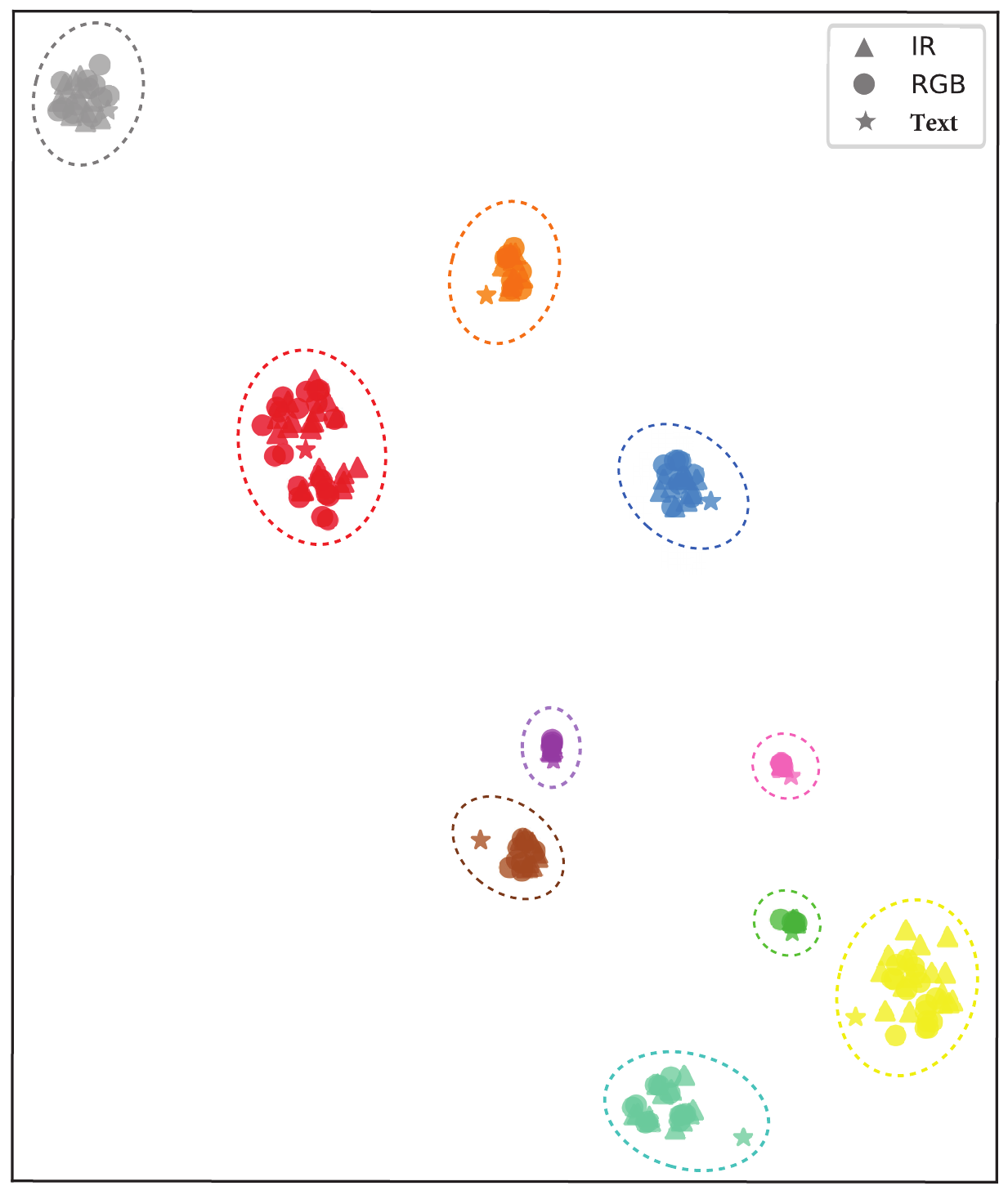}}
  \caption{T-SNE visualizations depict 10 randomly selected identities from the RegDB datasets. Here, different shapes represent different modalities, different colors denote distinct ids. Specifically, circles denote the RGB modality, triangles represent the IR modality and pentagrams represent the text modality.}
  \label{tsne}
\end{figure*}

\subsection{Ablation Studies}
\subsubsection{Ablation on Components} To further validate the effectiveness of each proposed component, we conduct an ablation study on the SYSU-MM01 and RegDB datasets, as shown in Table \ref{sysu_ablation} - \ref{regdb_ablation}. The analysis is performed by incrementally adding each component of our framework and evaluating the performance in terms of Rank-1 accuracy and mAP. Note that, the ``Base'' means that the pre-trained CLIP's image encoder is directly used to perform VI-ReID.
From the results on SYSU-MM01 dataset (Table \ref{sysu_ablation}), we observe that adding TSG improves performance over the baseline, indicating that learning semantically rich textual descriptions for visible images 
can enhance cross-modal alignment to some extent. Introducing IFE further improves the results, demonstrating the benefit of using textual guidance to optimize the shared encoder for extracting modality-shared 
features. When HSA is incorporated, which jointly optimizes the visual encoders and the textual features, a clear additional gain is achieved in both all-search and indoor-search settings. 
The full model achieves the best results, with 75.54\% Rank-1 and 72.55\% mAP on all-search, and 83.98\% Rank-1 and 85.99\% mAP on indoor-search.

Similar trends can be observed on RegDB dataset (Table \ref{regdb_ablation}), i.e., each component consistently contributes to performance improvement. Particularly, the joint optimization of HSA leads to a substantial performance boost in both visible-to-infrared and infrared-to-visible settings. These results clearly demonstrate that each component of our proposed framework is essential and complementary. The progressive integration of components leads to steady and substantial performance gains, verifying the effectiveness of our design.

\subsubsection{Ablation on Learnable Prompts} We also perform comparison experiments to verify the influence of learnable text tokens, and the experimental results are shown in Table \ref{learnable_token}. In this table, ``A photo of [class]'' method means we use a fixed text template as a semantic bridge to assist in cross-modal alignment. It can be observe that ``Learnable prompts (Token = 4)'' improves the Rank-1 accuracy and mAP by 2.67\% and 3.92\% under visible-to-infrared setting, and by 2.65\% and 4.44\% under infrared-to-visible setting, which substantiates the rationality of our motivation and affirms the effectiveness of the prompt learning. Furthermore, compared to the fixed template method, ``Learnable prompts'' can learn more detailed and essential semantic descriptions of pedestrians, rather than simple category labels. More detailed semantic descriptions help increase the inter-class distance between different pedestrians, thereby making VI-ReID easier.

Furthermore, we investigate the sensitivity of prompt length by evaluating different numbers of learnable tokens ($M$ = 2, 4, 6, and 8) on the RegDB dataset. From Table \ref{learnable_token}, we can see that as the number of learnable tokens increases, the performance first improves and reaches its best results at $M = 4$. When the number of tokens is further increased, the performance remains relatively stable without significant degradation, indicating that the proposed method is not overly sensitive to prompt length within a reasonable range.


\subsection{Parameter Analysis}
In the proposed CLIP4VI-ReID model, two hyperparameters are utilized, denoted as $\lambda_1$ and $\lambda_2$, to balance the importance of different loss terms throughout the entire training process. This section includes a parameter analysis for each hyperparameter to ascertain their optimal values. The corresponding results are presented in Fig. \ref{lambda1} and Fig. \ref{lambda2}.
In these figures, when determining the optimal value of $\lambda_1$, we initially set $\lambda_2$ to 0.05 and found that the model's performance was optimal when $\lambda_1$ took the value of 0.05. Therefore, we concluded that $\lambda_1 = 0.05$ is the optimal value. Subsequently, we fixed $\lambda_1 = 0.05$ and further explored the optimal value of $\lambda_2$, ultimately determining that the model performed best when $\lambda_2 = 0.05$. By employing these best values, we can effectively optimize the network weights of the model, thereby enhancing the model's overall performance in cross-modal ReID tasks.

\subsection{Complexity Analysis}
In order to evaluate the computational efficiency of the proposed method, we further report the parameter count, FLOPs, and training time compared with the baseline. The results are shown in Table \ref{time}. Notably, the CNN-based CLIP is used as our backbone, and the ``baseline'' refers to CLIP's image encoder. As shown by the experimental results, the proposed method slightly increases training time due to the adoption of a staged optimization strategy, yet achieves significant performance improvement with only acceptable additional overhead in terms of model parameters and training computational cost. This demonstrates that the proposed framework achieves a favorable trade-off between performance and efficiency. In addition, although the proposed method introduces additional optimization for the text branch during training, only the visual encoder is used during inference. Therefore, the inference-time computational complexity remains nearly identical to that of the baseline model, which verifies the high inference efficiency of our method.

\begin{table}[!t]
\renewcommand\arraystretch{1.06}

\begin{center}
\caption{Comparison of the training time and achieved performance on RegDB dataset.
}
\label{time}
\vspace{-3mm}
\resizebox{0.485\textwidth}{!}{ 
    \begin{tabular}{c|c| cc|cc| c c | c c }
    \hline
    \multirow{2}*{Methods} & \multirow{2}*{Training time} &\multicolumn{2}{c|}{Params/M} &\multicolumn{2}{c|}{FLOPs/G}  & \multicolumn{2}{c|}{Visible to Infrared} & \multicolumn{2}{c}{Infrared to Visible}  \\
    \cline{3-10}
   & & Training & Inference & Training & Inference & Rank-1 & mAP & Rank-1 & mAP  \\
    \hline 
    baseline  &5 hours  & 38.58 & 38.58 & 8.16 & 8.16 &86.84	&80.24	&84.76	&79.74\\ 
    \hline
    Ours      &11 hours & 79.38 & 38.58 & 11.90 & 8.16 & 94.51 & 88.55 & 92.28 & 86.58 \\
    \hline
    \end{tabular}
    }
\end{center}
\end{table}

\subsection{Visualization}
In order to more intuitively verify the discriminative ability of each stage, in Fig. \ref{tsne}, we perform qualitative analysis (i.e. 2D visualization) on RegDB dataset by using t-SNE \cite{van2008visualizing} method, we randomly select 10 identities from RegDB dataset, with each identity comprising 30 samples.
Fig. \ref{tsne}(a) shows that before training, features from the infrared and RGB modalities are distributed in different regions of the embedding space, exhibiting a significant modality gap and poor cross-modal identity alignment. 
Fig. \ref{tsne}(b) indicates that after applying TSG, the textual features extracted from RGB images are mostly clustered near the visible features, achieving preliminary alignment between visible images and text semantics.
Fig. \ref{tsne}(c) indicates that with text semantics as bridge, IFE eliminates the modality gap between RGB images and IR images. However, there are still significant overlaps between different categories, which limits the distinguishability of identities. In contrast, Fig. \ref{tsne}(d) shows that after the HSA stage, the CLIP4VI-ReID method enables each identity to be clearly and compactly clustered together. Meanwhile, it effectively aligns multi-modality features (including text features), significantly reduces modality differences, and improves discriminative ability.

\begin{figure}[t!]
	\centering
	\includegraphics[width=0.9\linewidth]{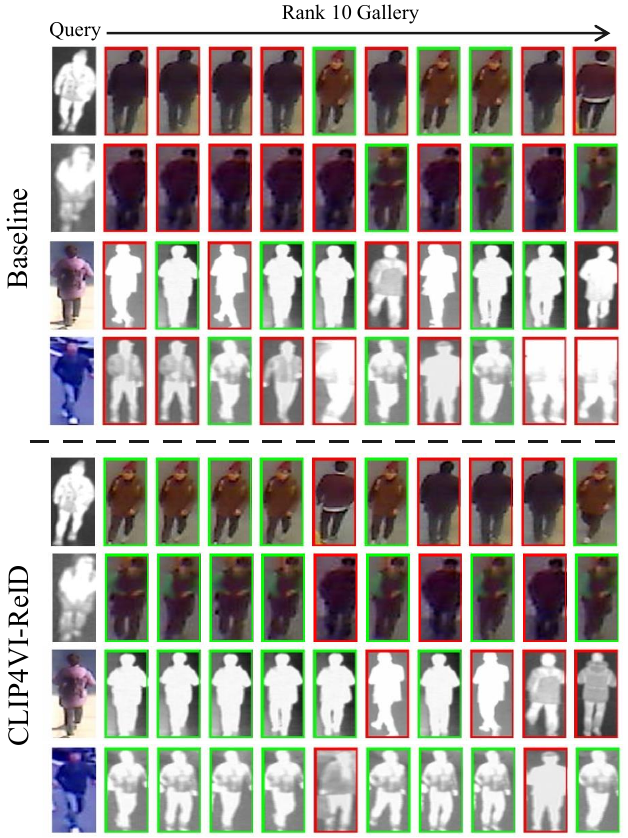} %
	\caption{Rank-10 retrieval outcomes from the baseline and our CLIP4VI-ReID.}
	\label{visual}
\end{figure}

To provide deeper insights into the effectiveness of the proposed CLIP4VI-ReID, we further conduct a qualitative comparison of retrieval results against the baseline on several representative image pairs from the RegDB dataset, as illustrated in Fig. \ref{visual}.  Here, the baseline represents the pre-trained CLIP model. In each example, the query images (first column) are from the IR (RGB) modality, while the gallery images (subsequent columns) are from the RGB (IR) modality. Correctly retrieved images that share the same identity with the query are highlighted with green bounding boxes, whereas incorrect retrievals are indicated with red bounding boxes. Overall, it can be observed that the proposed CLIP4VI-ReID consistently delivers improved retrieval performance, yielding a higher proportion of correct matches in the top-ranked positions compared to the baseline. This visually confirms the superior ability of our approach to bridge the cross-modality gap and enhance identity discriminability across IR and RGB domains.

\begin{figure}[t!]
\centering
\includegraphics[width=0.9\linewidth]{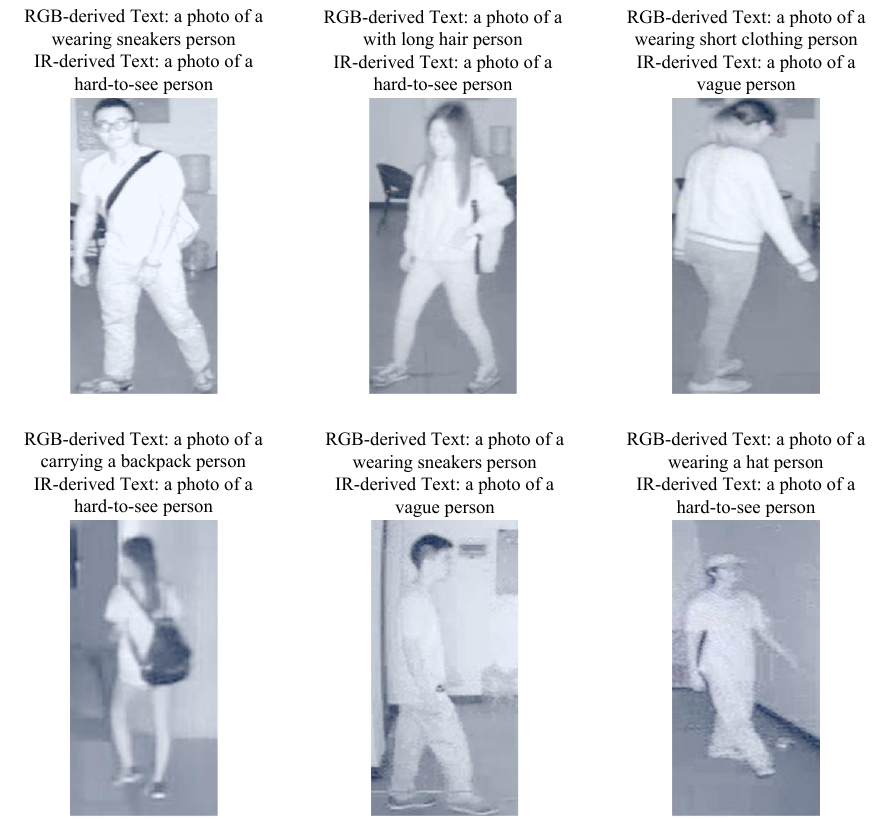} %
\caption{Failure cases of IR-derived text semantics. For each image, the top shows the text semantics derived from the IR image and RGB image. It can be observed that IR-derived semantics contain noisy or irrelevant information (e.g., “hard-to-see”, “vague”), whereas RGB-derived semantics capture ID-related attributes more accurately.}
\label{failure}
\end{figure}

Finally, to further verify the inherent limitations of directly generating text semantics from IR images (and thus justify the rationality of adopting RGB-derived semantics for IR alignment), we present representative failure cases in Fig. \ref{failure}. In this figure, for each pedestrian sample, we present the infrared (IR) image along with the text semantics generated from the corresponding visible (RGB) image (top) and the IR image (bottom). It can be observed that the IR-derived semantics contain noisy or irrelevant information (e.g., ``a hard-to-see person'', ``a vague person''), whereas the RGB-derived semantics capture identity-related attributes more accurately (e.g., ``a wearing sneakers person'', ``a carrying a backpack person''). This demonstrates that text semantics generated from RGB images are far more suitable for cross-modal alignment with IR images.

\section{Conclusion}
\label{sec:5}
In this paper, we propose the CLIP4VI-ReID method for visible-infrared person re-identification task, which utilizes the text semantics in CLIP as a bridge to progressively learn modality invariance features and enhance cross-modal alignment. Specifically, CLIP4VI-ReID contains three learning stages: text semantic generation, infrared feature embedding and high-level semantic alignment. The TSG stage first leverages the pre-trained CLIP text encoder to generate text semantics from visible images. Then, the IFE stage leverages the text semantics as a bridge to guide the learning of IR modality features and achieves visible-infrared cross-modal alignment to some extent. Finally, the HSA stage fine tunes the entire three-stream network to achieve finer higher-level semantic alignment among the three modalities and enhance the discriminative ability of modal shared semantics. A large number of experiments are performed on public VI-ReID datasets to demonstrate the effectiveness and superiority of the CLIP4VI-ReID method. 

Although the proposed CLIP4VI-ReID framework achieves promising performance on benchmark datasets, it still has several limitations that deserve discussion. First, our method generates modality-shared text semantics based on prompt learning, yet the text generation capacity of the prompt learning paradigm is limited, which restricts the richness of the learned semantics. In addition, the three-stage training pipeline leads to relatively high computational complexity. Second, ReID is a fine-grained visual task, while the CLIP model adopted in this paper is inherently coarse-grained, which presents limitations in fine-grained alignment and thus degrades cross-modal retrieval performance. Therefore, our future research will focus on designing text generation algorithms with richer semantic expressions and enhancing the fine-grained alignment capacity of CLIP, so as to more effectively guide the alignment of cross-modal visual features.


\bibliographystyle{IEEEtran}
\bibliography{IEEEabrv,data}

\begin{thebibliography}{10}
\providecommand{\url}[1]{#1}
\csname url@samestyle\endcsname
\providecommand{\newblock}{\relax}
\providecommand{\bibinfo}[2]{#2}
\providecommand{\BIBentrySTDinterwordspacing}{\spaceskip=0pt\relax}
\providecommand{\BIBentryALTinterwordstretchfactor}{4}
\providecommand{\BIBentryALTinterwordspacing}{\spaceskip=\fontdimen2\font plus
\BIBentryALTinterwordstretchfactor\fontdimen3\font minus
  \fontdimen4\font\relax}
\providecommand{\BIBforeignlanguage}[2]{{%
\expandafter\ifx\csname l@#1\endcsname\relax
\typeout{** WARNING: IEEEtran.bst: No hyphenation pattern has been}%
\typeout{** loaded for the language `#1'. Using the pattern for}%
\typeout{** the default language instead.}%
\else
\language=\csname l@#1\endcsname
\fi
#2}}
\providecommand{\BIBdecl}{\relax}
\BIBdecl

\bibitem{Zhu2025AE-Net}
S.~Zhu, Y.~Zhang, Y.~Liu, Y.~Feng, S.~Coleman, and D.~Kerr, ``Ae-net:
  Appearance-enriched neural network with foreground enhancement for person
  re-identification,'' \emph{IEEE Transactions on Emerging Topics in
  Computational Intelligence}, vol. Early Access, pp. 1--15, 2025.

\bibitem{ye2021deep}
M.~Ye, J.~Shen, G.~Lin, T.~Xiang, L.~Shao, and S.~C. Hoi, ``Deep learning for
  person re-identification: A survey and outlook,'' \emph{IEEE Transactions on
  Pattern Analysis and Machine Intelligence}, vol.~44, no.~6, pp. 2872--2893,
  2021.

\bibitem{Yan2024IsISILA}
S.~Yan, H.~Tang, L.~Zhang, and J.~Tang, ``Image-specific information
  suppression and implicit local alignment for text-based person search,''
  \emph{IEEE Transactions on Neural Networks and Learning Systems}, vol.~35,
  no.~12, pp. 17\,973--17\,986, 2024.

\bibitem{dong2023erasing}
N.~Dong, L.~Zhang, S.~Yan, H.~Tang, and J.~Tang, ``Erasing, transforming, and
  noising defense network for occluded person re-identification,'' \emph{IEEE
  Transactions on Circuits and Systems for Video Technology}, vol.~34, no.~6,
  pp. 4458--4472, 2023.

\bibitem{zhang2022person}
M.~Zhang, Y.~Xiao, F.~Xiong, S.~Li, Z.~Cao, Z.~Fang, and J.~T. Zhou, ``Person
  re-identification with hierarchical discriminative spatial aggregation,''
  \emph{IEEE Transactions on Information Forensics and Security}, vol.~17, pp.
  516--530, 2022.

\bibitem{Singh2025TROPE}
D.~Singh, J.~Mathew, M.~Agarwal, and M.~Govind, ``Trope: Triplet-guided feature
  refinement for person re-identification,'' \emph{IEEE Transactions on
  Emerging Topics in Computational Intelligence}, vol.~9, no.~1, pp. 706--716,
  2025.

\bibitem{Zhang2022MFEA}
L.~Zhang, K.~Li, and Y.~Qi, ``Person re-identification with multi-features
  based on evolutionary algorithm,'' \emph{IEEE Transactions on Emerging Topics
  in Computational Intelligence}, vol.~6, no.~3, pp. 509--518, 2022.

\bibitem{zhang2022dual}
Y.~Zhang, Y.~Kang, S.~Zhao, and J.~Shen, ``Dual-semantic consistency learning
  for visible-infrared person re-identification,'' \emph{IEEE Transactions on
  Information Forensics and Security}, vol.~18, pp. 1554--1565, 2022.

\bibitem{lu2023tri}
Z.~Lu, R.~Lin, and H.~Hu, ``Tri-level modality-information disentanglement for
  visible-infrared person re-identification,'' \emph{IEEE Transactions on
  Multimedia}, vol.~26, pp. 2700--2714, 2023.

\bibitem{zhang2022fmcnet}
Q.~Zhang, C.~Lai, J.~Liu, N.~Huang, and J.~Han, ``Fmcnet: Feature-level
  modality compensation for visible-infrared person re-identification,'' in
  \emph{Proceedings of the IEEE/CVF Conference on Computer Vision and Pattern
  Recognition}, 2022, pp. 7349--7358.

\bibitem{cui2024dma}
Z.~Cui, J.~Zhou, and Y.~Peng, ``Dma: Dual modality-aware alignment for
  visible-infrared person re-identification,'' \emph{IEEE Transactions on
  Information Forensics and Security}, vol.~19, pp. 2696--2708, 2024.

\bibitem{lu2023learning}
H.~Lu, X.~Zou, and P.~Zhang, ``Learning progressive modality-shared
  transformers for effective visible-infrared person re-identification,'' in
  \emph{Proceedings of the AAAI Conference on Artificial Intelligence},
  vol.~37, no.~2, 2023, pp. 1835--1843.

\bibitem{liu2022revisiting}
J.~Liu, J.~Wang, N.~Huang, Q.~Zhang, and J.~Han, ``Revisiting modality-specific
  feature compensation for visible-infrared person re-identification,''
  \emph{IEEE Transactions on Circuits and Systems for Video Technology},
  vol.~32, no.~10, pp. 7226--7240, 2022.

\bibitem{qiu2024high}
L.~Qiu, S.~Chen, Y.~Yan, J.-H. Xue, D.-H. Wang, and S.~Zhu, ``High-order
  structure based middle-feature learning for visible-infrared person
  re-identification,'' in \emph{Proceedings of the AAAI Conference on
  Artificial Intelligence}, vol.~38, no.~5, 2024, pp. 4596--4604.

\bibitem{fang2023visible}
X.~Fang, Y.~Yang, and Y.~Fu, ``Visible-infrared person re-identification via
  semantic alignment and affinity inference,'' in \emph{Proceedings of the
  IEEE/CVF International Conference on Computer Vision}, 2023, pp.
  11\,270--11\,279.

\bibitem{zhao2022spatial}
J.~Zhao, H.~Wang, Y.~Zhou, R.~Yao, S.~Chen, and A.~El~Saddik, ``Spatial-channel
  enhanced transformer for visible-infrared person re-identification,''
  \emph{IEEE Transactions on Multimedia}, vol.~25, pp. 3668--3680, 2022.

\bibitem{hu2022adversarial}
W.~Hu, B.~Liu, H.~Zeng, Y.~Hou, and H.~Hu, ``Adversarial decoupling and
  modality-invariant representation learning for visible-infrared person
  re-identification,'' \emph{IEEE Transactions on Circuits and Systems for
  Video Technology}, vol.~32, no.~8, pp. 5095--5109, 2022.

\bibitem{Chan2025DFHFL}
S.~Chan, W.~Meng, Z.~Li, J.~Hu, and X.~Zhou, ``Diverse-feature hierarchical
  fusion learning for visible-infrared person re-identification,'' \emph{IEEE
  Transactions on Emerging Topics in Computational Intelligence}, vol. Early
  Access, pp. 1--11, 2025.

\bibitem{wei2023dual}
Z.~Wei, X.~Yang, N.~Wang, and X.~Gao, ``Dual-adversarial representation
  disentanglement for visible infrared person re-identification,'' \emph{IEEE
  Transactions on Information Forensics and Security}, vol.~19, pp. 2186--2200,
  2023.

\bibitem{chen2023identity}
X.~Chen, X.~Zheng, and X.~Lu, ``Identity feature disentanglement for
  visible-infrared person re-identification,'' \emph{ACM Transactions on
  Multimedia Computing, Communications and Applications}, vol.~19, no.~6, pp.
  1--20, 2023.

\bibitem{liang2023cross}
T.~Liang, Y.~Jin, W.~Liu, and Y.~Li, ``Cross-modality transformer with modality
  mining for visible-infrared person re-identification,'' \emph{IEEE
  Transactions on Multimedia}, vol.~25, pp. 8432--8444, 2023.

\bibitem{feng2022visible}
Y.~Feng, J.~Yu, F.~Chen, Y.~Ji, F.~Wu, S.~Liu, and X.-Y. Jing,
  ``Visible-infrared person re-identification via cross-modality interaction
  transformer,'' \emph{IEEE Transactions on Multimedia}, vol.~25, pp.
  7647--7659, 2022.

\bibitem{liu2020enhancing}
H.~Liu, J.~Cheng, W.~Wang, Y.~Su, and H.~Bai, ``Enhancing the discriminative
  feature learning for visible-thermal cross-modality person
  re-identification,'' \emph{Neurocomputing}, vol. 398, pp. 11--19, 2020.

\bibitem{xiang2019cross}
X.~Xiang, N.~Lv, Z.~Yu, M.~Zhai, and A.~El~Saddik, ``Cross-modality person
  re-identification based on dual-path multi-branch network,'' \emph{IEEE
  Sensors Journal}, vol.~19, no.~23, pp. 11\,706--11\,713, 2019.

\bibitem{yu2025clip}
\BIBentryALTinterwordspacing
X.~Yu, N.~Dong, L.~Zhu, H.~Peng, and D.~Tao, ``Clip-driven semantic discovery
  network for visible-infrared person re-identification,'' \emph{IEEE
  Transactions on Multimedia}, 2024. [Online]. Available:
  \url{https://arxiv.org/abs/2401.05806}
\BIBentrySTDinterwordspacing

\bibitem{wang2025learning}
J.~Wang, X.~Gao, S.~Niu, H.~Zhao, G.~Feng, and J.~Lin, ``Learning
  discriminative features via deep metric learning for video-based person
  re-identification,'' \emph{Expert Systems with Applications}, vol. 286, p.
  128123, 2025.

\bibitem{wang2025DIRL}
J.~Wang, X.~Gao, S.~Niu, H.~Zhao, and G.~Feng, ``Dirl: Learning discriminative
  id-related representations for video visible-infrared person re-id,''
  \emph{ACM Transactions on Multimedia Computing, Communications and
  Applications}, vol.~21, no.~8, pp. 238:1--16, Jun. 2025.

\bibitem{li2023clip}
S.~Li, L.~Sun, and Q.~Li, ``Clip-reid: exploiting vision-language model for
  image re-identification without concrete text labels,'' in \emph{Proceedings
  of the AAAI Conference on Artificial Intelligence}, vol.~37, no.~1, 2023, pp.
  1405--1413.

\bibitem{Tang2025CGSKTLM}
H.~Tang, S.~He, and J.~Qin, ``Connecting giants: Synergistic knowledge transfer
  of large multimodal models for few-shot learning,'' in \emph{Proceedings of
  the Thirty-Fourth International Joint Conference on Artificial Intelligence,
  {IJCAI-25}}, J.~Kwok, Ed.\hskip 1em plus 0.5em minus 0.4em\relax
  International Joint Conferences on Artificial Intelligence Organization, 8
  2025, pp. 6227--6235, main Track.

\bibitem{radford2021learning}
A.~Radford, J.~W. Kim, C.~Hallacy, A.~Ramesh, G.~Goh, S.~Agarwal, G.~Sastry,
  A.~Askell, P.~Mishkin, J.~Clark \emph{et~al.}, ``Learning transferable visual
  models from natural language supervision,'' in \emph{International Conference
  on Machine Learning}.\hskip 1em plus 0.5em minus 0.4em\relax PmLR, 2021, pp.
  8748--8763.

\bibitem{bahng2022exploring}
H.~Bahng, A.~Jahanian, S.~Sankaranarayanan, and P.~Isola, ``Exploring visual
  prompts for adapting large-scale models,'' \emph{arXiv preprint
  arXiv:2203.17274}, 2022.

\bibitem{zhou2022learning}
K.~Zhou, J.~Yang, C.~C. Loy, and Z.~Liu, ``Learning to prompt for
  vision-language models,'' \emph{International Journal of Computer Vision},
  vol. 130, no.~9, pp. 2337--2348, 2022.

\bibitem{yu2024tf}
C.~Yu, X.~Liu, Y.~Wang, P.~Zhang, and H.~Lu, ``Tf-clip: Learning text-free clip
  for video-based person re-identification,'' in \emph{Proceedings of the AAAI
  Conference on Artificial Intelligence}, vol.~38, no.~7, 2024, pp. 6764--6772.

\bibitem{Hu2025CLIPMC}
G.~Hu, Y.~Lv, J.~Zhang, Q.~Wu, and Z.~Wen, ``Clip-based modality compensation
  for visible-infrared image re-identification,'' \emph{IEEE Transactions on
  Multimedia}, vol.~27, pp. 2112--2126, 2025.

\bibitem{cao2025irgpt}
Z.~Cao, J.~Zhang, and R.~Zhang, ``Irgpt: Understanding real-world infrared
  image with bi-cross-modal curriculum on large-scale benchmark,'' in
  \emph{Proceedings of the IEEE/CVF International Conference on Computer
  Vision}, 2025, pp. 166--176.

\bibitem{Wang2025NPSSL}
J.~Wang, J.~Wen, W.~Ding, C.~Yu, X.~Zhu, and Z.~Wang, ``Noise perception
  self-supervised learning for unsupervised person re-identification,''
  \emph{IEEE Transactions on Emerging Topics in Computational Intelligence},
  vol. Early Access, pp. 1--11, 2025.

\bibitem{he2021transreid}
S.~He, H.~Luo, P.~Wang, F.~Wang, H.~Li, and W.~Jiang, ``Transreid:
  Transformer-based object re-identification,'' in \emph{Proceedings of the
  IEEE/CVF International Conference on Computer Vision}, 2021, pp.
  15\,013--15\,022.

\bibitem{gao2024part}
S.~Gao, C.~Yu, P.~Zhang, and H.~Lu, ``Part representation learning with
  teacher-student decoder for occluded person re-identification,'' in
  \emph{ICASSP 2024-2024 IEEE International Conference on Acoustics, Speech and
  Signal Processing (ICASSP)}.\hskip 1em plus 0.5em minus 0.4em\relax IEEE,
  2024, pp. 2660--2664.

\bibitem{zhu2019progressive}
Z.~Zhu, T.~Huang, B.~Shi, M.~Yu, B.~Wang, and X.~Bai, ``Progressive pose
  attention transfer for person image generation,'' in \emph{Proceedings of the
  IEEE/CVF Conference on Computer Vision and Pattern Recognition}, 2019, pp.
  2347--2356.

\bibitem{hermans2017defense}
A.~Hermans, L.~Beyer, and B.~Leibe, ``In defense of the triplet loss for person
  re-identification,'' \emph{arXiv preprint arXiv:1703.07737}, 2017.

\bibitem{luo2019strong}
H.~Luo, W.~Jiang, Y.~Gu, F.~Liu, X.~Liao, S.~Lai, and J.~Gu, ``A strong
  baseline and batch normalization neck for deep person re-identification,''
  \emph{IEEE Transactions on Multimedia}, vol.~22, no.~10, pp. 2597--2609,
  2019.

\bibitem{ye2016person}
M.~Ye, C.~Liang, Y.~Yu, Z.~Wang, Q.~Leng, C.~Xiao, J.~Chen, and R.~Hu, ``Person
  reidentification via ranking aggregation of similarity pulling and
  dissimilarity pushing,'' \emph{IEEE Transactions on Multimedia}, vol.~18,
  no.~12, pp. 2553--2566, 2016.

\bibitem{sarfraz2018pose}
M.~S. Sarfraz, A.~Schumann, A.~Eberle, and R.~Stiefelhagen, ``A pose-sensitive
  embedding for person re-identification with expanded cross neighborhood
  re-ranking,'' in \emph{Proceedings of the IEEE Conference on Computer Vision
  and Pattern Recognition}, 2018, pp. 420--429.

\bibitem{chen2022structure}
C.~Chen, M.~Ye, M.~Qi, J.~Wu, J.~Jiang, and C.-W. Lin, ``Structure-aware
  positional transformer for visible-infrared person re-identification,''
  \emph{IEEE Transactions on Image Processing}, vol.~31, pp. 2352--2364, 2022.

\bibitem{li2020infrared}
D.~Li, X.~Wei, X.~Hong, and Y.~Gong, ``Infrared-visible cross-modal person
  re-identification with an x modality,'' in \emph{Proceedings of the AAAI
  Conference on Artificial Intelligence}, vol.~34, no.~04, 2020, pp.
  4610--4617.

\bibitem{Tang2025DaC}
H.~Tang, Z.~Li, D.~Zhang, S.~He, and J.~Tang, ``Divide-and-conquer: Confluent
  triple-flow network for rgb-t salient object detection,'' \emph{IEEE
  Transactions on Pattern Analysis and Machine Intelligence}, vol.~47, no.~3,
  pp. 1958--1974, 2025.

\bibitem{Tang2022LAgPF}
H.~Tang, C.~Yuan, Z.~Li, and J.~Tang, ``Learning attention-guided pyramidal
  features for few-shot fine-grained recognition,'' \emph{Pattern Recognition},
  vol. 130, p. 108792, 2022.

\bibitem{wu2017rgb}
A.~Wu, W.-S. Zheng, H.-X. Yu, S.~Gong, and J.~Lai, ``Rgb-infrared
  cross-modality person re-identification,'' in \emph{Proceedings of the IEEE
  International Conference on Computer Vision}, 2017, pp. 5380--5389.

\bibitem{nguyen2017person}
D.~T. Nguyen, H.~G. Hong, K.~W. Kim, and K.~R. Park, ``Person recognition
  system based on a combination of body images from visible light and thermal
  cameras,'' \emph{Sensors}, vol.~17, no.~3, p. 605, 2017.

\bibitem{wang2019rgb}
G.~Wang, T.~Zhang, J.~Cheng, S.~Liu, Y.~Yang, and Z.~Hou, ``Rgb-infrared
  cross-modality person re-identification via joint pixel and feature
  alignment,'' in \emph{Proceedings of the IEEE/CVF International Conference on
  Computer Vision}, 2019, pp. 3623--3632.

\bibitem{ye2020dynamic}
M.~Ye, J.~Shen, D.~J.~Crandall, L.~Shao, and J.~Luo, ``Dynamic dual-attentive
  aggregation learning for visible-infrared person re-identification,'' in
  \emph{Computer Vision--ECCV 2020: 16th European Conference, Glasgow, UK,
  August 23--28, 2020, Proceedings, Part XVII 16}.\hskip 1em plus 0.5em minus
  0.4em\relax Springer, 2020, pp. 229--247.

\bibitem{wu2021discover}
Q.~Wu, P.~Dai, J.~Chen, C.-W. Lin, Y.~Wu, F.~Huang, B.~Zhong, and R.~Ji,
  ``Discover cross-modality nuances for visible-infrared person
  re-identification,'' in \emph{Proceedings of the IEEE/CVF Conference on
  Computer Vision and Pattern Recognition}, 2021, pp. 4330--4339.

\bibitem{liu2021sfanet}
H.~Liu, S.~Ma, D.~Xia, and S.~Li, ``Sfanet: A spectrum-aware feature
  augmentation network for visible-infrared person reidentification,''
  \emph{IEEE Transactions on Neural Networks and Learning Systems}, vol.~34,
  no.~4, pp. 1958--1971, 2021.

\bibitem{huang2022modality}
Z.~Huang, J.~Liu, L.~Li, K.~Zheng, and Z.-J. Zha, ``Modality-adaptive mixup and
  invariant decomposition for rgb-infrared person re-identification,'' in
  \emph{Proceedings of the AAAI Conference on Artificial Intelligence},
  vol.~36, no.~1, 2022, pp. 1034--1042.

\bibitem{yu2023toplight}
H.~Yu, X.~Cheng, and W.~Peng, ``Toplight: lightweight neural networks with
  task-oriented pretraining for visible-infrared recognition,'' in
  \emph{Proceedings of the IEEE/CVF Conference on Computer Vision and Pattern
  Recognition}, 2023, pp. 3541--3550.

\bibitem{wu2024enhancing}
R.~Wu, B.~Jiao, W.~Wang, M.~Liu, and P.~Wang, ``Enhancing visible-infrared
  person re-identification with modality-and instance-aware visual prompt
  learning,'' in \emph{Proceedings of the 2024 International Conference on
  Multimedia Retrieval}, 2024, pp. 579--588.

\bibitem{lu2024disentangling}
Z.~Lu, R.~Lin, and H.~Hu, ``Disentangling modality and posture factors:
  Memory-attention and orthogonal decomposition for visible-infrared person
  re-identification,'' \emph{IEEE Transactions on Neural Networks and Learning
  Systems}, vol.~36, no.~3, pp. 5494--5508, 2025.

\bibitem{liu2024cross}
M.~Liu, Z.~Zhang, Y.~Bian, X.~Wang, Y.~Sun, B.~Zhang, and Y.~Wang,
  ``Cross-modality semantic consistency learning for visible-infrared person
  re-identification,'' \emph{IEEE Transactions on Multimedia}, 2024.

\bibitem{zhang2023diverse}
Y.~Zhang and H.~Wang, ``Diverse embedding expansion network and low-light
  cross-modality benchmark for visible-infrared person re-identification,'' in
  \emph{Proceedings of the IEEE/CVF Conference on Computer Vision and Pattern
  Recognition}, 2023, pp. 2153--2162.

\bibitem{kingma2014adam}
D.~P. Kingma and J.~Ba, ``Adam: A method for stochastic optimization,''
  \emph{arXiv preprint arXiv:1412.6980}, 2014.

\bibitem{van2008visualizing}
L.~Van~der Maaten and G.~Hinton, ``Visualizing data using t-sne.''
  \emph{Journal of Machine Learning Research}, vol.~9, no.~11, 2008.

\end{thebibliography}

\vspace{11pt}

\begin{IEEEbiography}[{\includegraphics[width=1in,height=1.25in,clip,keepaspectratio]{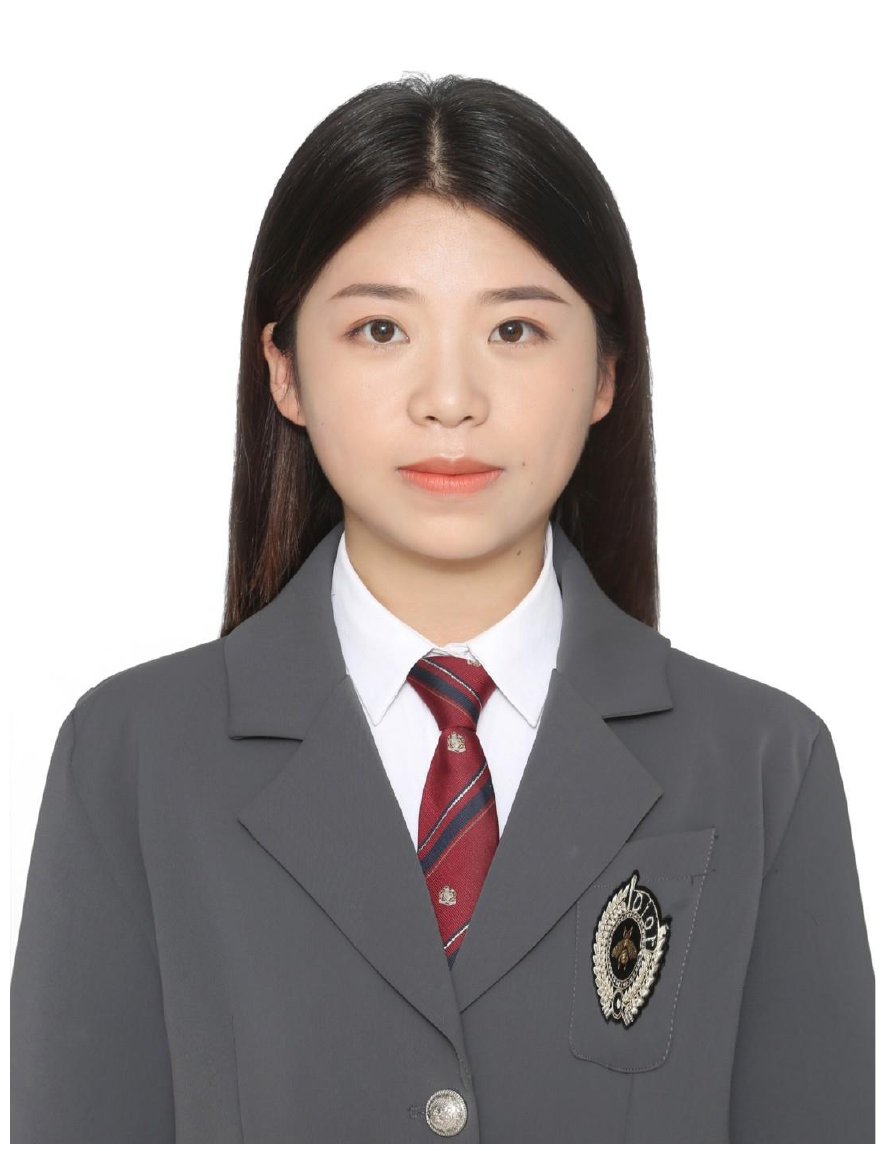}}]{Xiaomei Yang}
received her B.S. Degree from the School of Computer Science at the Qufu Normal University, in 2023.

Now she is currently studying for a master's Degree in the school of Information Science and Engineering, University of Jinan. Her research interests include person re-identification, video analysis and multi-model machine learning.
\end{IEEEbiography}
\vspace{-1.5em}

\begin{IEEEbiography}[{\includegraphics[width=1in,height=1.25in,clip,keepaspectratio]{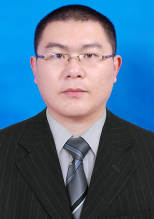}}]{Xizhan Gao}
received the Ph.D. degree in control science and engineering from Nanjing University of Science and Technology, Nanjing, China, in 2019.

He is currently an associate professor of the School of Information Science and Engineering, University of Jinan. His research interests include pattern recognition, computer vision and video processing.
\end{IEEEbiography}
\vspace{-1.5em}

\begin{IEEEbiography}[{\includegraphics[width=1in,height=1.25in,clip,keepaspectratio]{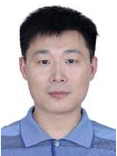}}]{Sijie Niu}
received B.S. and Ph.D. Degrees from the School of Computer science at Liaocheng University and Nanjing University of Science and Technology in 2007 and 2016, respectively. He was a visiting scholar at Stanford University in 2014. He was a Post-doctoral with medical image analysis, UNC in 2021. He is currently a professor in the School of Information Science and Engineering, University of Jinan, China. His research interests include Pattern recognition, machine learning, image processing, and medical image analysis.
\end{IEEEbiography}
\vspace{-1.5em}

\begin{IEEEbiography}[{\includegraphics[width=1in,height=1.25in,clip,keepaspectratio]{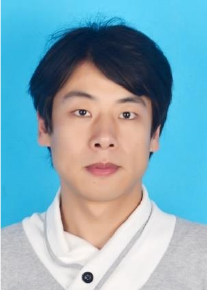}}]{Fa Zhu}
(Member, IEEE) received his Ph.D. in Control Science and Engineering from the School of Computer Science and Engineering, Nanjing University of Science and Technology, Nanjing, P.R. China in 2019. From 2016 to 2018, he was a Visiting Student with the Centre for Artificial Intelligence (CAI) and the Faculty of Engineering and Information Technology, University of Technology Sydney, Ultimo, NSW, Australia. Most of his researches have published on high prestigious journals, such as IEEE TNNLS, IEEE TIFS, IEEE TVT, IEEE TCE, IEEE IOTJ, IEEE JBHI, PR, Inf Sci, ESWA, EAAI, ASOC, NEURO etc. He is/was associate/guest editor of IEEE TCE, DCN, CAEE, HCIS, CMC etc. His current research interests include pattern recognition and machine learning, anomaly detection, Internet of Things etc.
\end{IEEEbiography}
\vspace{-1.5em}

\begin{IEEEbiography}[{\includegraphics[width=1in,height=1.25in,clip,keepaspectratio]{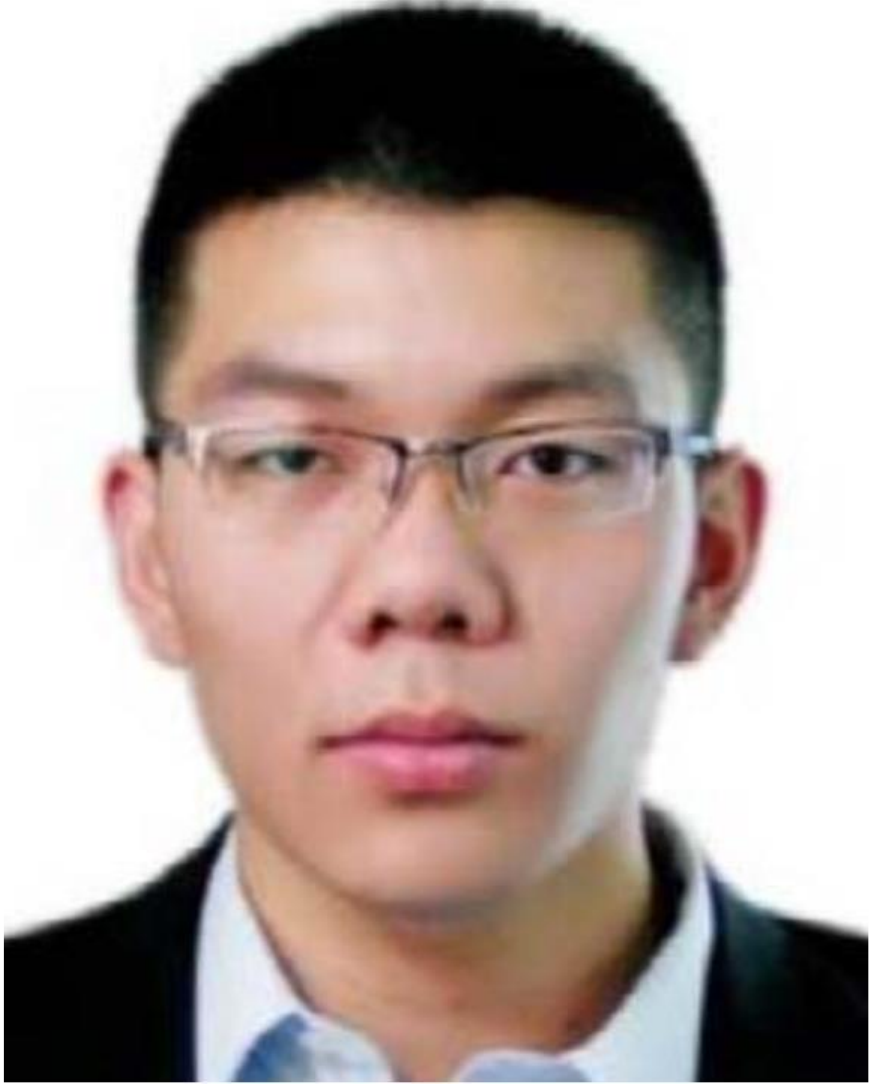}}]{Guang Feng}
received the B.E. degree in electronic information engineering from Qingdao University in 2015, and the M.E. degree in signal and information processing from the University of Jinan in 2018. He received the Ph.D. degree in School of Information and Communication Engineering, Dalian University of Technology in 2022. He is currently an associate professor of the School of Information Science and Engineering, University of Jinan. His research interests include saliency detection and referring expression comprehension.
\end{IEEEbiography}
\vspace{-1.5em}

\begin{IEEEbiography}[{\includegraphics[width=1in,height=1.25in,clip,keepaspectratio]{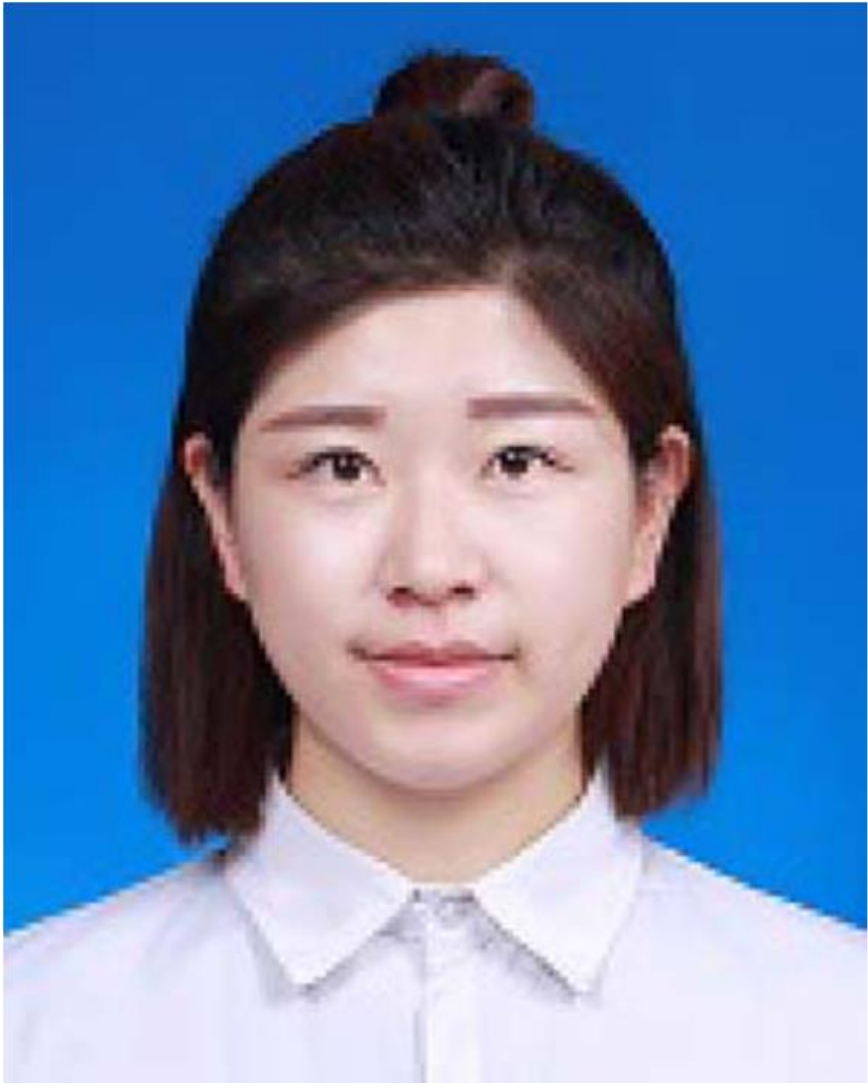}}]{Xiaofeng Qu}
received the Ph.D. degree from School of Information Science and Engineering, Shandong Normal University in 2024. 

She is currently a lecturer of the School of Information Science and Engineering, University of Jinan. Her research interests include computer vision, deep learning, and person re-identification.
\end{IEEEbiography}
\vspace{-1.5em}

\begin{IEEEbiography}[{\includegraphics[width=1in,height=1.25in,clip,keepaspectratio]{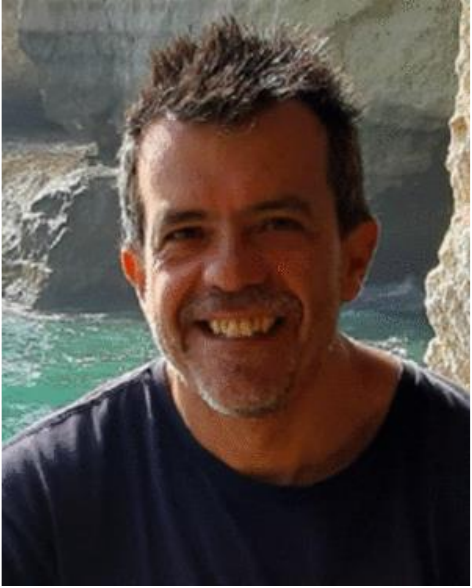}}]{David Camacho}
(Senior Member, IEEE) received the Ph.D. degree (with Honors) in Computer Xcience from Universidad Carlos III de Madrid, in 2001. He is currently a Full Professor with Computer Systems Engineering Department, Universidad Politécnica de Madrid (UPM), Madrid, Spain, and the Head of the Applied Intelligence and Data Analysis research Group, UPM. He has authored or coauthored more than 300 journals, books, and conference papers. His research interests include machine learning (clustering/deep learning), computational intelligence (evolutionary computation, swarm intelligence), social network analysis, fake news and disinformation analysis. He has participated/led more than 60 research projects (Spanish and European: H2020, DG Justice, ISFP, and Erasmus+), related to the design and application of artificial intelligence methods for data mining and optimization for problems emerging in industrial scenarios, aeronautics, aerospace engineering, cybercrime/cyber intelligence, social networks applications, or video games among others. Contact him at: david.camacho@upm.es.
\end{IEEEbiography}

\end{document}